# Discovering Semantic Granularities in Images through Clustering of Pretrained Visual Representations

Ying Dai

Iwate Prefectural University

dai@iwate-pu.ac.jp

## Abstract

Semantic granularity, which refers to the levels at which visual representations organize image content into meaningful entities, is fundamental to image understanding. However, whether pretrained visual representations inherently encode semantic structures at multiple granularities remains poorly understood. This paper presents a representation-driven approach for discovering semantically meaningful image regions at different granularities through clustering of pretrained visual representations without task-specific training. Specifically, we concatenate feature maps from appropriately deep layers of a pretrained visual model and apply a singular value decomposition (SVD)-based projection to derive a pixel-wise principal representation map (PRM) tensor. The resulting PRM tensor captures latent semantic structures embedded in pretrained visual representations, making them accessible for subsequent semantic clustering. We then introduce the semantic continuity ratio (SCR) and analyze the signal-to-noise ratio (SNR) of SCR profiles to assess the suitability of PRM tensors for semantic granularity discovery. Our results indicate that PRM tensors exhibiting a higher proportion of relatively high SNR values over an initial range of cluster numbers tend to yield more semantically meaningful regions across multiple granularities. Furthermore, once a suitable PRM tensor is identified through SNR-profile analysis, pixel-wise hierarchical clustering enables the level of semantic granularity to be controlled by varying the number of clusters. We evaluate the proposed approach on the COCO-2017 and VOC-2012 validation datasets using Hungarian-matched intersection over union (HIoU). The results suggest competitive segmentation performance among the compared methods, supporting the feasibility of the proposed approach.

## 1. Introduction

Visual understanding requires organizing an image into meaningful entities at different levels of abstraction. A scene may be interpreted as a collection of objects and stuffs, while an object may further be decomposed into parts or finer semantic components. This multi-level organization, referred to here as semantic granularities, is fundamental to visual perception because the appropriate level of interpretation depends on the visual context and the task. Although deep visual models are trained for recognition or representation learning, their intermediate features often preserve richer spatial and semantic information than is required by their training objectives. This raises a fundamental question: do pretrained visual representations inherently encode multiple levels of semantically coherent organization, and can these levels be discovered without task-specific training?

Unsupervised image segmentation has advanced through graph-based clustering and pretrained visual representations. Jiao et al. [1] integrated high-level features with superpixels and graph clustering, but their approach may be limited by superpixel quality and handcrafted cues. Current studies investigate semantic information emerging in pretrained representations. Zadaianchuk et al. [2] exploited self-supervised object-centric representations. DiffCut [3] combined diffusion features with recursive normalized cuts for zero-shot semantic segmentation, and DeepCut++ [4] integrated feature fusion and diffusion-based graph refinement. DiffSeg [5] proposed to utilize the self-attention layers in stable diffusion models for unsupervised segmentation because the pre-trained stable diffusion model has learned inherent concepts of objects within its attention layers. MaskCut [6] proposed a simple approach of Cut-and-Learn for training unsupervised object detection and segmentation models. They leverage the property of self-supervised models to discover objects without supervision. Although these methods demonstrate the potential of pretrained representations, their primary objective is generally to obtain a segmentation of a particular image, rather than to investigate whether a pretrained representation itself contains multiple intrinsic levels of semantic organization.

More recent methods have advanced unsupervised semantic segmentation by learning dense features specifically for clustering. PiCIE [7], for example, incorporated invariance and equivariance constraints to learn pixel-wise representations that support semantic clustering in complex, multi-object scenes. STEGO [8] instead distilled semantic correspondences from self-supervised features into a segmentation model, demonstrating that pretrained representations can contain dense semantic information even without human annotations. These methods have shown that pretrained visual features can support surprisingly strong semantic segmentation. Nevertheless, they generally optimize the representation or segmentation model toward a desired segmentation objective. Consequently, the granularity of the resulting grouping is largely treated as a property of the task or model, rather than as a phenomenon to be discovered from the representation itself.

The issue of granularity becomes particularly important in hierarchical segmentation. Ke et al. [9] recognized that unsupervised semantic segmentation is inherently ambiguous because meaningful groupings can occur at different levels of granularity. Their Hierarchical Segment Grouping therefore learns representations that support consistent grouping across multiple levels. Similarly, recent hierarchy-agnostic approaches seek to construct semantic region hierarchies from latent representations without imposing a predefined semantic hierarchy [10]. These studies establish that visual representations can support hierarchical organization and that multiple grouping levels are meaningful. However, they primarily formulate the problem as learning or constructing a hierarchical segmentation system. The more basic question of whether multiple semantic granularities are already latent in a pretrained visual representation, and how such granularities can be identified directly from its intrinsic organization, remains less explored.

Self-supervised visual representation learning provides further evidence that deep features can encode semantic structure beyond their explicit training objectives. DINO showed that self-supervised vision Transformer features can exhibit information about object and semantic segmentation, despite being learned without semantic labels [11]. TokenCut exploited these self-supervised transformer features for unsupervised object discovery through graph-based normalized cuts [12]. Such results suggest that pretrained representations may organize image content according to meaningful visual entities. However, these approaches typically focus on discovering a salient object or a predefined type of segmentation, rather than systematically examining how the representation behaves as the desired number of regions changes. Thus, they provide limited insight into whether a single representation can naturally support multiple semantic granularities ranging from coarse objects to finer structures.

Recently, Kim et al. [13] leveraged causal inference and concept-wise self-supervised learning to discover semantic regions at multiple granularities, but depend on heuristic concept merging. Yu et al. [14] extend SAM 2 [15] with self-supervised granularity control, yet rely on a pretrained foundation model rather than independently discovering semantic granularity from pretrained visual representations.

Dai [16] found that pixel-wise latent feature representations derived from EfficientNet-B0 [17], a convolutional neural network (CNN), can yield high quality ingredient segmentation in food images without task-specific training. However, the discovery of different meaningful regions at multiple granularity levels, such as dish-level and ingredient-level regions, remains unexplored. These observations motivate a different perspective: we view this study as an investigation into how latent semantic organization embedded in pretrained visual representations can be captured and exploited. If a representation encodes semantic structures at multiple levels, varying the number of clusters should reveal distinct partitions that remain internally coherent and externally distinguishable across different granularities. Conversely, a representation dominated by low-level appearance similarities may produce visually homogeneous regions that do not correspond to meaningful semantic entities.

In this paper, we propose a representation-driven approach to discovering semantically meaningful image regions at different granularities through clustering of pretrained visual representations without task-specific training. we concatenate feature maps from appropriately deep layers of a pretrained visual model and apply a singular value decomposition (SVD)-based projection to derive a pixel-wise principal representation map (PRM) tensor. The resulting PRM tensor captures latent semantic structures embedded in pretrained visual representations, making them accessible for subsequent semantic clustering. By varying the number of clusters, the resulting partitions can reveal different semantic regions at multiple granularity levels.

We then introduce the semantic continuity ratio (SCR), based on the silhouette principle, to quantify the pixel-wise clustering behavior of PRM tensor at different candidate granularities. Furthermore, we analyze the signal-to-noise ratio (SNR), defined as the ratio of the mean to the sample standard deviation of SCR values over an initial range of cluster numbers to assess the suitability of the PRM tensor for semantic granularity discovery. The results indicate that PRM tensor exhibiting a higher proportion of relatively high SNR values over an initial range of cluster numbers tend to yield more semantically meaningful regions across multiple granularities. Furthermore, once a suitable PRM tensor is identified through SNR-profile analysis, pixel-wise hierarchical clustering enables the level of semantic granularity to be controlled by varying the number of clusters, with the pixel-wise representation dimensionality of the PRM tensor set equal to the number of clusters.

The main contributions of this paper are summarized as follows:

1. We propose PRM tensors that capture latent semantic structures embedded in pretrained visual representations, making these structures accessible for subsequent semantic clustering.
2. We introduce SCR and SCR profile-based SNR analysis to assess the suitability of PRM tensors for semantic granularity discovery without task-specific training. The qualitative results indicate that EfficientNet-B0 provides particularly suitable representations for this purpose, as assessed by the SNR profiles.
3. We evaluate the proposed approach on the COCO-2017 and VOC-2012 validation datasets using Hungarian-matched intersection over union (HIoU). The results show competitive segmentation performance among the compared methods, supporting the feasibility of the proposed method.

## 2. Methodology

### 2.1. Framework Overview

The proposed framework discovers semantic granularities directly from pretrained visual representations without task-specific training. It consists of four stages: (1) constructing a PRM tensor from pretrained visual representations; (2) performing pixel-wise clustering of the PRM tensor at different numbers of clusters; (3) assessing the suitability of PRMs for semantic granularity discovery using semantic continuity ratio (SCR) profiles and their signal-to-noise ratios (SNRs); and (4) externally evaluating the resulting segmentations using ground-truth (GT) semantic masks.

The central idea is that a representation capable of encoding meaningful semantic structures should produce identifiable regions at multiple clustering granularities. Accordingly, rather than evaluating a representation at a single cluster number, we investigate its clustering behavior across an initial range of cluster numbers and examine whether its SCR profile indicates favorable clustering behavior.

Figure 1 illustrates the construction of PRM tensor from an input image and the subsequent segmentation results through hierarchical clustering.

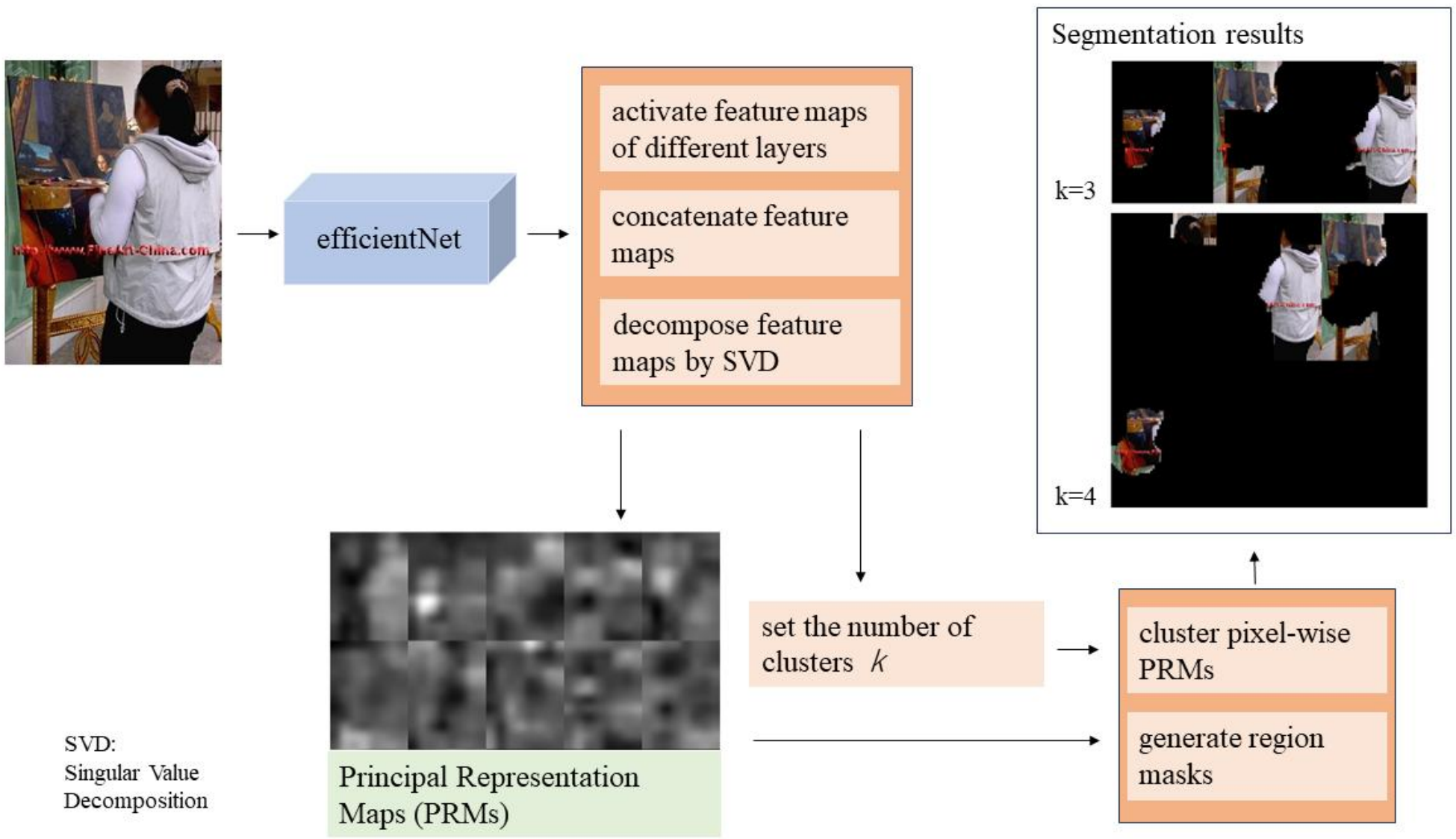

Figure 1 Overview of constructing PRM tensor and clustering

## 2.2. Construction of principal representation map tensor

Inspired by the method proposed in [16], we introduce how to construct PRM tensor using the pretrained visual representations. Let $F^{(l)}$ denote the feature maps extracted from the $l$-th layer of a pretrained visual model for a given image. Since deeper layers generally encode more abstract semantic information, we concatenate feature maps from appropriately selected deep layers to construct a pixel-wise representation that integrates complementary semantic information.

First, the selected feature maps are spatially aligned to a common resolution of $H * W$. where $H$ and $W$ denote its spatial height and width, respectively. The aligned feature maps are then concatenated along the channel dimension to produce a unified feature map tensor:

$$F_{cat} = concat(F^{(l_1)}, F^{(l_2)}, \dots, F^{(l_m)}) \quad (1)$$

where $concat(\cdot)$ denote the concatenation operation, $F^{(l_m)}$ denotes the feature maps of $l_m$-th network layers and $m$ denotes the index of the selected layer.

Let $C$ denote the total number of channels across the selected layers. The resulting feature tensor has dimensions $H * W * C$.

To obtain coherent semantic information, SVD is applied to $F_{cat}$ to construct PRM tensor. First, $F_{cat}$ is reshaped into a matrix of $HW * C$. Then, Z-score standardization of this matrix is performed so that each dimension has zero mean and unit variance; the resulting standardized matrix is denoted as Z. SVD is then applied to Z:

$$Z = U\Sigma V^T \quad (2)$$

where U is an orthogonal matrix whose columns are the left singular vectors; V is an orthogonal matrix whose columns are the right singular vectors; and Σ is a diagonal matrix containing the singular values.

By retaining the first $n$ singular values and their corresponding singular vectors, the projection scores of $Z$ are computed as

$$Z_{SVD} = U_n \Sigma_n \quad (3)$$

where $Z_{SVD}$ denotes a matrix of projection scores of $F_{cat}$ in the new coordinate system, expressed along the principal directions. Furthermore, $Z_{SVD}$ is reshaped into a spatial tensor with the dimensions $H * W * n$. Then, the first $n_p$ maps of the tensor are extracted to construct the PRM tensor. Each pixel is represented by a $n_p$-dimensional feature vector that integrates semantic information from the selected feature maps.

Figure 2 shows the first 10 maps of PRM tensor derived from EfficientNet-B0, along with the first 10 channel-wise feature maps with the highest mean values extracted from the selected layers for a given image that is selected from ADE20K validation dataset.

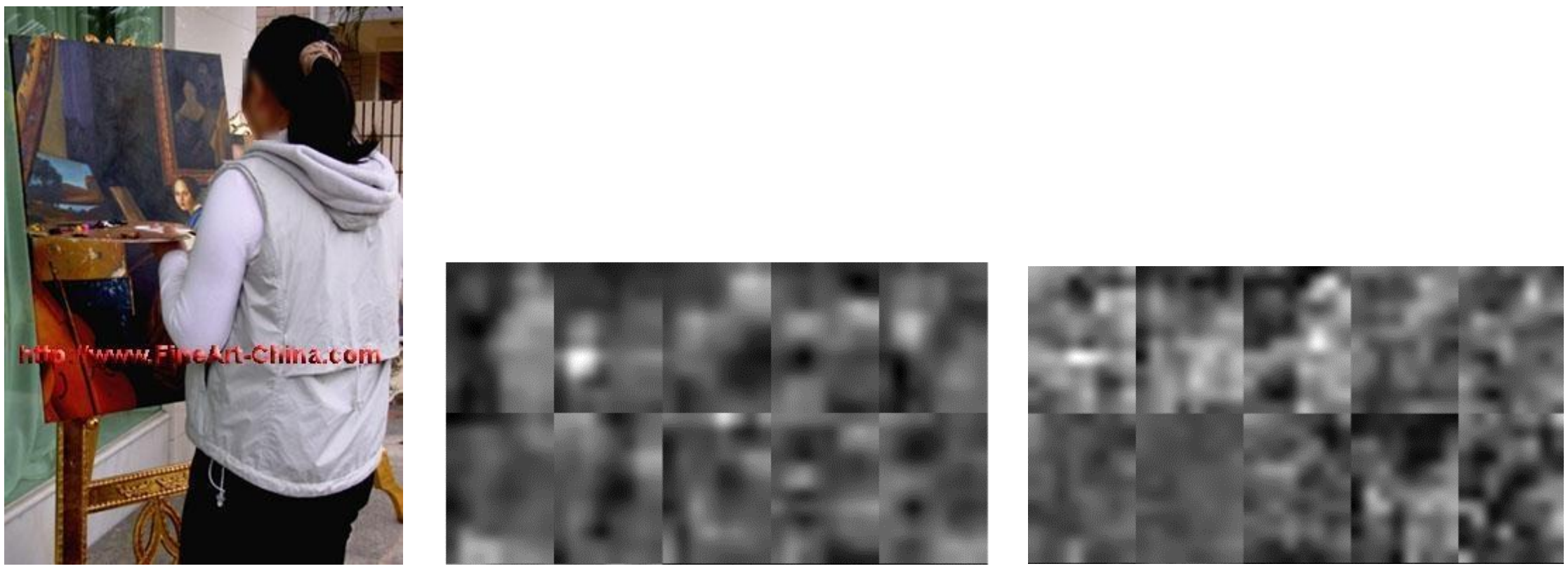

(a) Original image (b) PRMs (c) Feature maps with highest means

Figure 2 PRMs and maximal feature maps

As shown in Figure 2, the first three maps exhibit clear and meaningful spatial organization. The first map highlights the person region, the second emphasizes the red-colored region and the paint tray, and the third focuses on the head region. In contrast, the original channel-wise feature maps do not exhibit similarly meaningful spatial organization. These observations indicate that the maps of PRM tensor capture the latent semantic structures embedded in the pretrained representations, making them accessible to subsequent semantic clustering.

### 2.3. Clustering at multiple granularities

Given the PRM tensor, each pixel is represented by an $n_p$-dimensional feature vector. Since each PRM map appears to capture a distinct semantic-like spatial structure, we set the number of PRMs equal to the number of clusters, i.e., $n_p = k$, where $k$ denotes the number of clusters. This design aligns the representation dimensionality with the target clustering granularity. Hierarchical clustering is then applied to partition the image into spatially distributed regions based on feature similarity. For each candidate value of $k$, clustering produces a label map. By varying $k$, the framework generates a collection of segmentation maps representing different levels of granularity. Generally, smaller values of $k$ produce coarser partitions, whereas larger values permit finer partitions. However, increasing the cluster number does not necessarily improve semantic meaningfulness: a finer partition may separate meaningful entities, but it may also divide semantically coherent regions into unnecessary fragments.

Therefore, the objective is not simply to optimize the number of clusters or select a single segmentation with the highest clustering score. Instead, we investigate whether the representation produces semantically coherent partitions over a range of cluster numbers.

### 2.4. Semantic continuity ratio

To characterize clustering behavior across different granularities, we introduce the semantic continuity ratio (SCR), motivated by the silhouette principle. Silhouette score (SS) measures how similar an individual data point is to its own cluster (cohesion) compared to other clusters (separation). The score ranges from -1 to 1, where a higher score indicates the better-defined cluster. Accordingly, SCR is defined by the following expression (4).

$$SCR = \frac{the\ number\ of\ points\ with\ SS > T}{the\ number\ of\ all\ points} \tag{4}$$

It means the proportion of points which have the SS larger than a threshold T to all points. We use T=0.7 to identify pixels exhibiting relatively strong within-cluster cohesion and between-cluster separation. A higher SCR indicates better clustering quality for points. The SCR is used to characterize the coherence of PRM-based clustering at each candidate cluster number. Its purpose is to assess whether the resulting regions exhibit favorable within-region coherence and between-region separation.

For a representation tensor, evaluating SCR across a sequence of cluster numbers yields an SCR profile:

$$SCRP = \{SCR(k) \mid k \in K\} \tag{5}$$

where SCR(k) denotes the SCR value for the representation at cluster number *k*, and *K* denotes the initial range of candidate cluster numbers.

Analyzing the complete profile, rather than an isolated SCR value, makes it possible to investigate whether PRM tensor maintains favorable clustering characteristics across multiple granularities. Importantly, SCR evaluates the clustering structure in the learned feature space; it does not require ground-truth semantic labels to identify candidate granularities.

We therefore introduce the signal-to-noise ratio (SNR), defined as the ratio of the mean SCR to its sample standard deviation (std) over an initial range of cluster numbers. The SNR is calculated using the equation (6).

$$SNR(K) = \frac{mean_{k<K}(SCR(k))}{std_{k<K}(SCR(k))} \tag{6}$$

where *mean(SCR(k)* represents the average cluster coherence of the clustering results across the selected cluster numbers, and *std(SCR(k))* represents the variability of cluster coherence across those cluster numbers. Thus, $SNR(K)$ characterizes both the average magnitude and the variability of SCR over the selected range. SNR is regarded as a SCR profile-based indicator of the consistency and magnitude of cluster coherence across multiple clustering granularities. A higher SNR indicates a more favorable balance between SCR magnitude and variation. In particular, a relatively high SNR suggests that the PRM tensor maintains relatively high and consistent SCR values over this range, indicating its suitability for discovering semantically meaningful regions at multiple granularities.

Moreover, rather than relying on a single SNR value, we examine the proportion of relatively high SNR values across different initial ranges of cluster numbers. This enables us to compare representations in terms of the consistency of their clustering behavior and investigate the relationship between favorable SNR profiles and the ability to yield meaningful regions across multiple granularities. It should be noted that no fixed threshold is defined for high SNR values, as their magnitudes can vary considerably across different images.

The resulting analysis provides a representation-level indicator for assessing the suitability of PRM tensors for semantic granularity discovery without requiring semantic supervision. We conjecture that A higher proportion of relatively high SNR values across the evaluated range of cluster numbers appears to indicate that the corresponding PRM tensor is more capable of capturing semantically meaningful regions at different granularities and producing high quality partitions. However, this indicator should be regarded as useful but not sufficient, since visually similar regions may also exhibit high SNR values.

For the image shown in Figure 2, Figure 3 shows its SNR profiles as a function of the initial range of cluster number for PRM tensor derived from EfficientNet-B0. The legend entry Layer A indicates that the PRM tensor are derived from the feature maps of layer 286. Layers A+B indicates that the PRM tensor is constructed by concatenating the feature maps of layers 286 and 269. Similarly, Layers A+B+C indicates that the PRM tensor are constructed by concatenating the feature maps of layers 286, 269, and 251. The blue curve represents the SNR profile for Layer A, and the other curves correspond to the respective layer configurations indicated in the legend. The starting point of the initial range is set to K=3, since the range corresponding to K=2 contains only a single data point, making the sample standard deviation undefined.

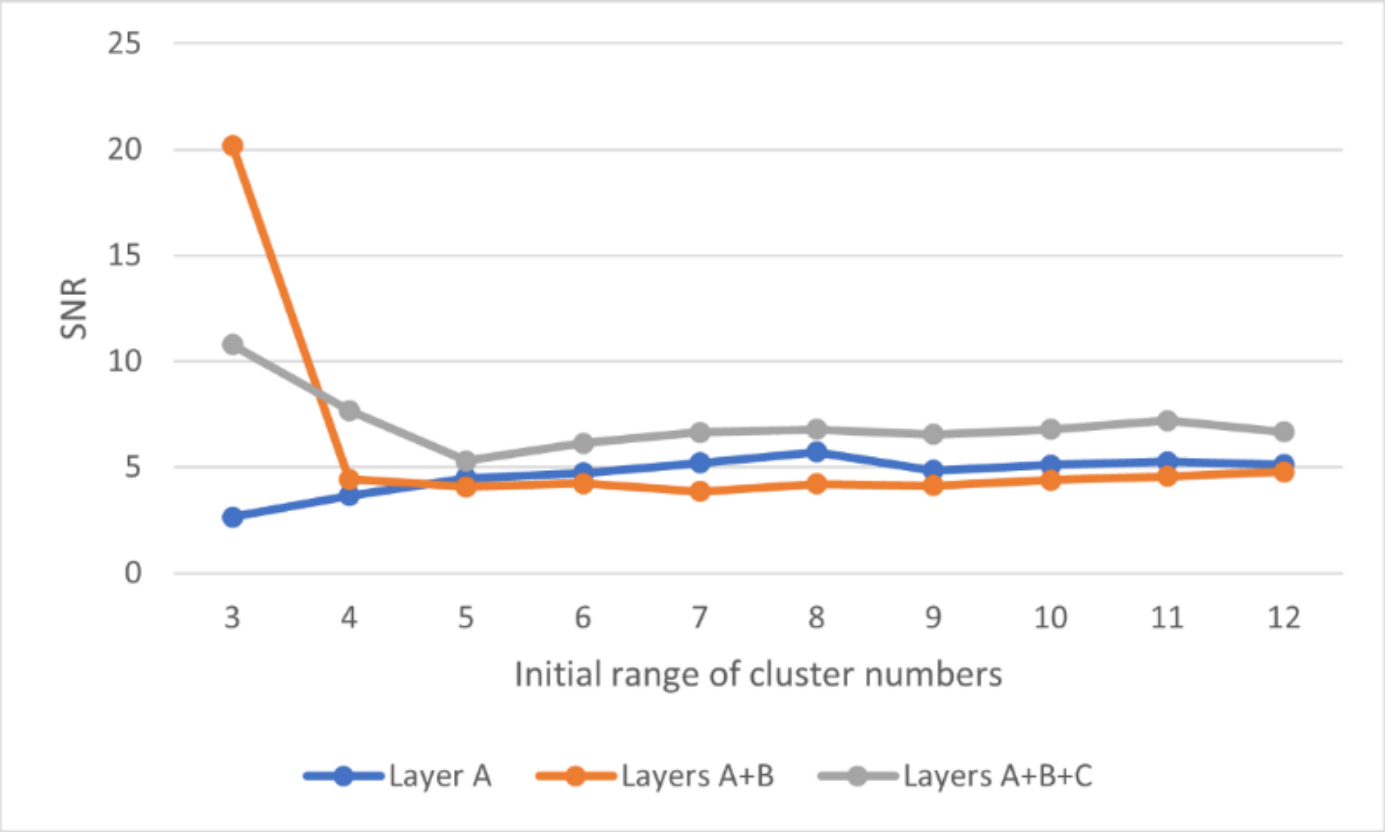

Figure 3 SNR profiles

As shown in Figure 3, Layer A exhibits low SNR values. Layers A+B achieve a high SNR over the initial range up to K=3, but the SNR decreases sharply as the range expands. In contrast, Layers A+B+C maintain comparatively high SNR values over the initial ranges up to K=3 and K=4. Although the SNR values decrease sharply as the range expands further, they remain comparatively high and stable thereafter. This indicates that Layers A+B+C exhibit a higher proportion of relatively high SNR values across the evaluated ranges. These results suggest that Layers A+B+C are more suitable for producing meaningful partitions across multiple levels of granularity, whereas Layer A appears less effective at capturing meaningful regions.

In order to support this conjecture, we show their first 5 PRMs in Figure 4.

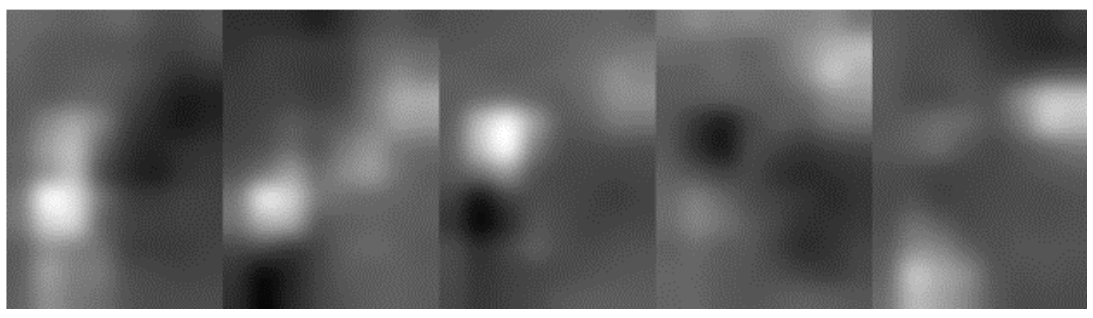

(a) Layer A

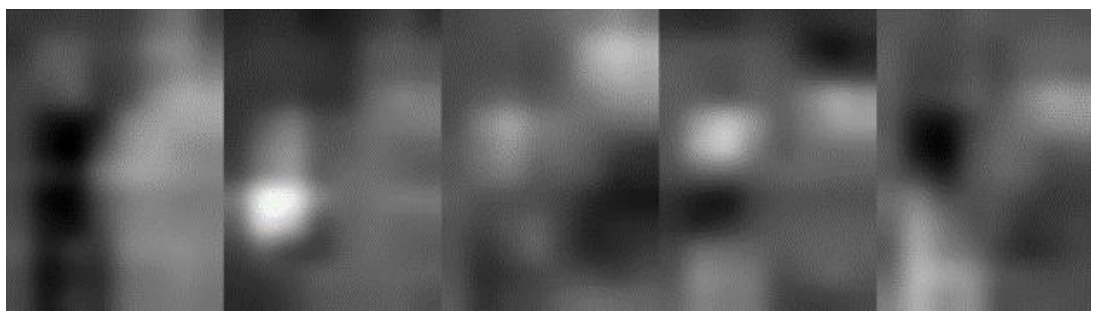

(b) Layers A+B

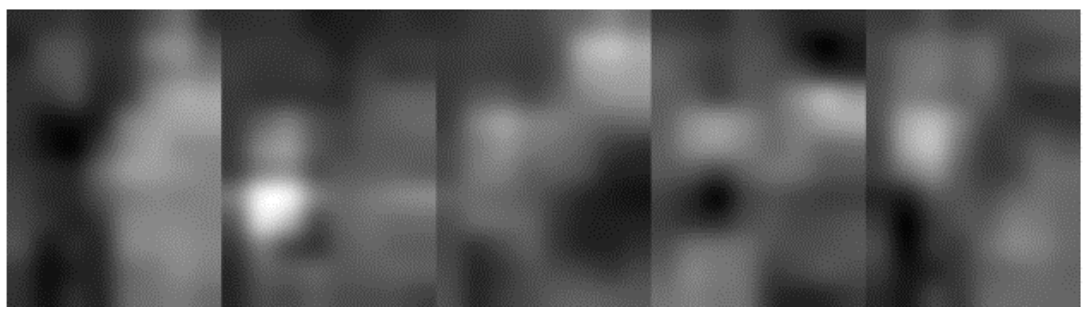

(c) Layers A+B+C

Figure 4 PRMs of different layers

As shown in Figure 4, for Layers A+B+C, each of the five maps highlights a meaningful semantic-like spatial organization, whereas for Layers A+B, only the first two maps clearly highlight meaningful regions. In contrast, for Layer A, none of the maps clearly highlights the person region. These observations illustrate a correspondence between the SNR profiles and the semantic-like spatial organization exhibited by the PRM tensor. Specifically, PRM tensor containing a greater number of maps that exhibit meaningful semantic-like spatial organization tend to be associated with more favorable SNR profiles.

Moreover, we show the corresponding segmentation results in Figure 5. Upper line are the results when k=3, and lower line are the results when k=4.

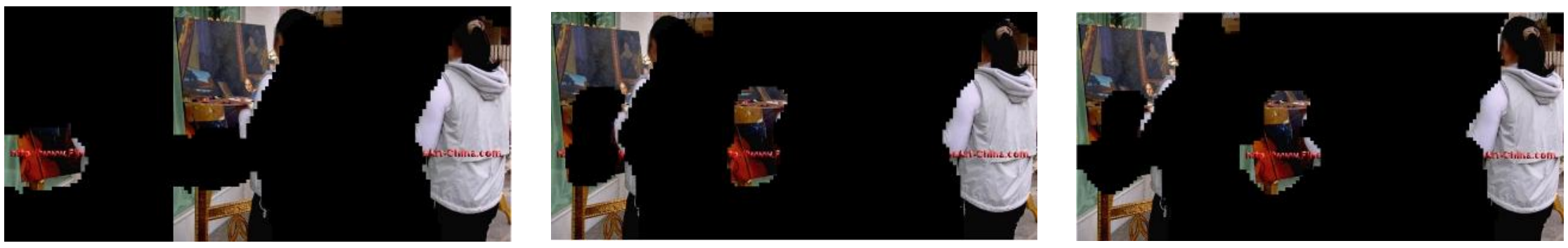

(a) Layer A (b) Layers A+B (c) Layers A+B+C

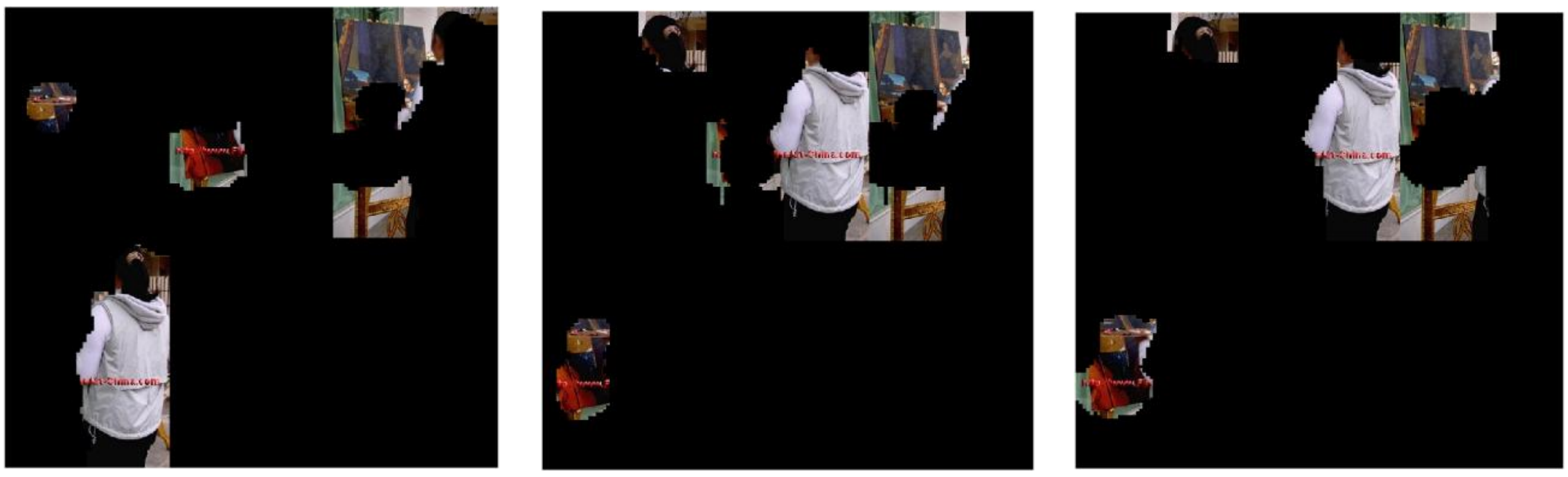

(d) Layer A (e) Layers A+B (f) Layers A+B+C

Figure 5 Segmentation results

As shown in Figure 5, when k=3, the person is segmented in all three cases. However, in the results obtained using Layer A and Layer A+B, a portion of the person is incorrectly assigned to the painting segment. In the case of Layer A, the paint tray is not identified as a separate region.

When k=4, the head and body are separated in the results obtained using Layer A+B and Layer A+B+C. However, with Layer A+B, a portion of the wall is incorrectly assigned to the body segment, whereas with Layer A, a portion of the person is incorrectly assigned to the painting segment. Overall, the PRM tensor derived from Layer A+B+C produce the best segmentation results.

These observations suggest a correspondence between SNR profiles and the quality of the resulting semantic partitions. The qualitative results support the use of SNR profiles as an indicator for assessing the suitability of PRM tensors for semantic granularity discovery. PRM tensors exhibiting a higher proportion of relatively high SNR values over an initial range of cluster numbers tend to yield more semantically meaningful regions across multiple granularities.

## 3. Experimental validation and analysis

In this section, we conduct experiments to evaluate the effectiveness of PRM tensors and validate the role of SCR profiles in assessing the suitability of PRM tensors for semantic granularity discovery.

Specifically, we address three questions: (1) how do PRM tensor derived from different pretrained models and feature-layer configurations differ in their semantic structures? (2) Are more favorable SNR profiles of SCR profiles associated with better segmentation quality? (3) Can SCR-profile analysis help identify the suitability of PRM tensor for semantic granularity discovery without task-specific training?

### 3.1. Experimental setup

The pretrained visual representations are derived from four pretrained visual models: EfficientNet-B0 [17], GoogleNet [18], ResNet-50 [19], and SAM 2 [15]. EfficientNet-B0, GoogLeNet and ResNet-50 are pretrained on ImageNet classification and SAM 2 on mask annotations; no training or fine-tuning is performed on VOC-2012 or COCO-2017.

For models that provide multiple feature maps from different network layers, we evaluate selected layer configurations. For EfficientNet-B0, the feature maps of layers 286, 269, and 251 are selected for concatenation; for GoogleNet, the feature maps of layers 138, 135, and 131 are selected; for ResNet-50, the feature maps of layers 173, 169 and 166 are selected.

For each applicable configuration, the extracted feature maps are spatially aligned and concatenated to form a combined representation. Each map has a spatial resolution of [64, round (64 × W/H)], where W and H denote the width and height of the input image, respectively. SVD is then applied to obtain PRM tensor. For SAM 2, the specific features used to construct PRM tensor are determined by its feature extraction procedure.

We then apply pixel-wise hierarchical clustering to PRM tensors on the COCO-2017 and VOC-2012 validation datasets using two settings: an adaptively determined number of clusters and the number of ground-truth (GT) masks. In the adaptive setting, the number of clusters is automatically determined based on the first slowdown point in the singular value spectrum. This setting allows us to evaluate the proposed approach without requiring prior knowledge of the number of GT masks. We evaluate the resulting partitions against the ground-truth (GT) masks using Hungarian Intersection-over-Union (HIoU). Specifically, we first compute pairwise IoU values between predicted clusters and GT masks using the Jaccard similarity. The Hungarian algorithm is then applied to determine the one-to-one matching that maximizes the total IoU. Finally, HIoU is calculated as the average IoU over the matched pairs, with unmatched clusters or GT masks contributing zero.

We conduct experiments using MATLAB 2026b on a Dell Alienware laptop equipped with a 12GB RTX 4090 Laptop GPU and a 64GB RAM.

We employ MATLAB's built-in linkage function to perform hierarchical clustering using the Ward linkage method with Euclidean distance. The resulting hierarchical cluster tree is then cut at different numbers of clusters to obtain partitions at multiple granularities.

### 3.2. Comparison of PRM tensor across pretrained models

As an example, we first visualize the PRM tensor of the image shown in Figure 2, derived from EfficientNet-B0, GoogleNet, ResNet-50, and SAM 2, respectively. The purpose is to examine whether different pretrained architectures produce representations with similarly meaningful semantic-like spatial organization. Figure 6 shows their first 10 PRMs, respectively.

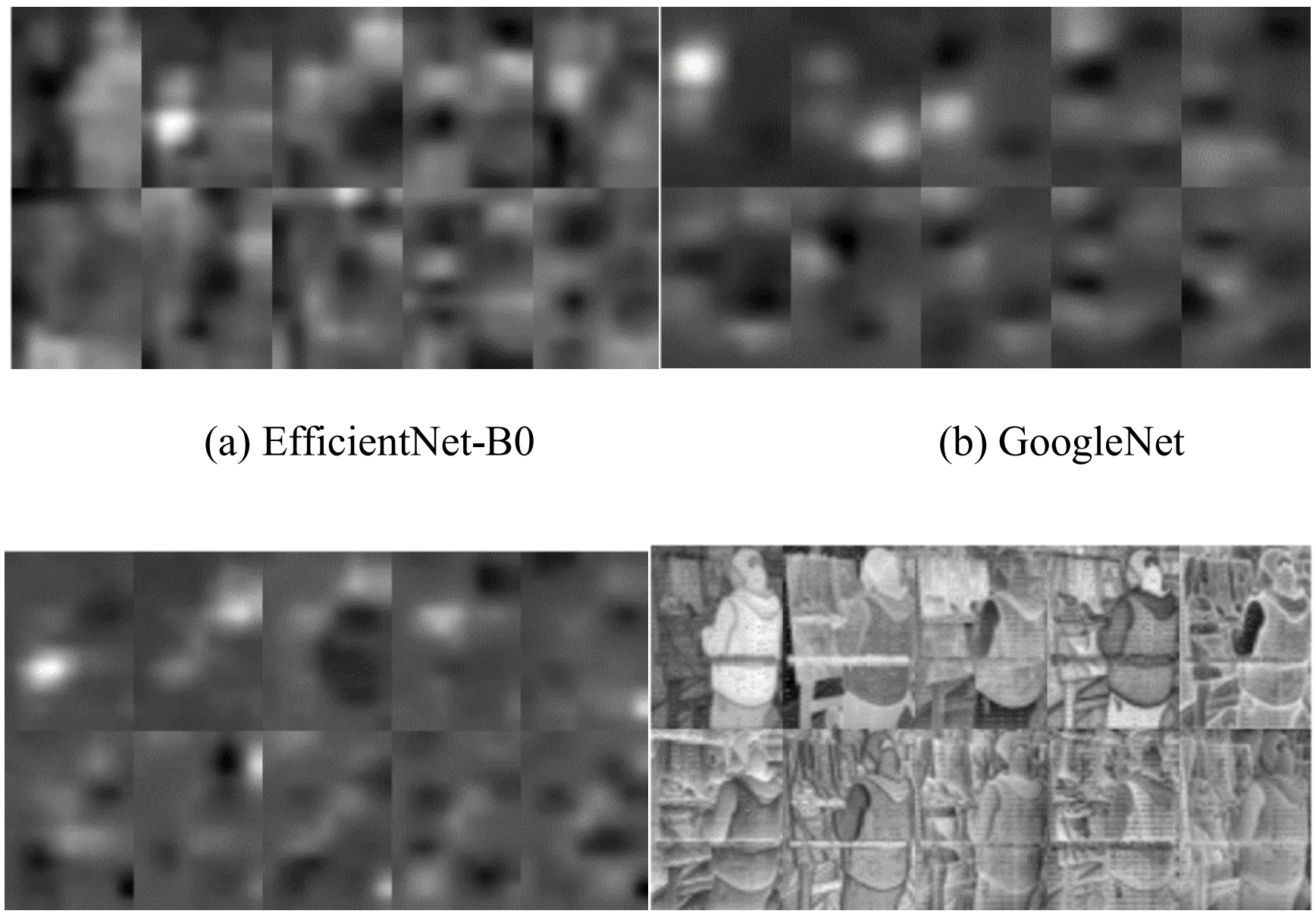

(a) EfficientNet-B0 (b) GoogleNet

(c) ResNet-50 (d) SAM 2

Figure 6 PRMs derived from four pretrained models

As shown in Figure 6, the first five PRMs derived from EfficientNet-B0 highlight multiple potentially meaningful semantic structures, including a person, a paint tray, and a head. The first PRM derived from GoogLeNet appears to emphasize a single meaningful region corresponding to the paint tray, whereas the first two PRMs derived from ResNet-50 appear to highlight two distinct regions: a red-colored region and a head region, respectively. In contrast, the PRMs derived from SAM 2 primarily emphasize local details and entity boundaries, suggesting a limited ability to reveal meaningful regions across multiple semantic granularities. These qualitative observations suggest that the PRMs derived from EfficientNet-B0 exhibit richer semantic-like spatial organizations, potentially making them more suitable for discovering semantically meaningful regions at multiple granularities through clustering.

Figure 7 shows the segmentation results obtained by applying pixel-wise hierarchical clustering to the first three PRMs, with the number of clusters set to three.

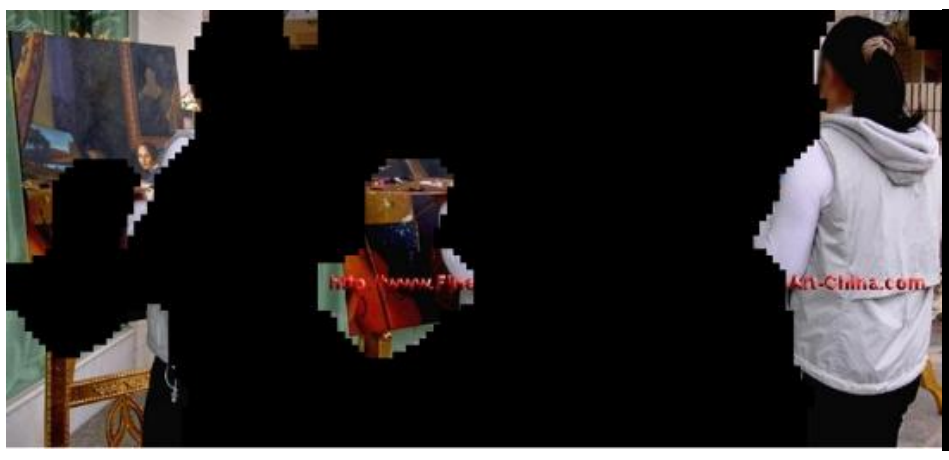

(a) EfficientNet-B0

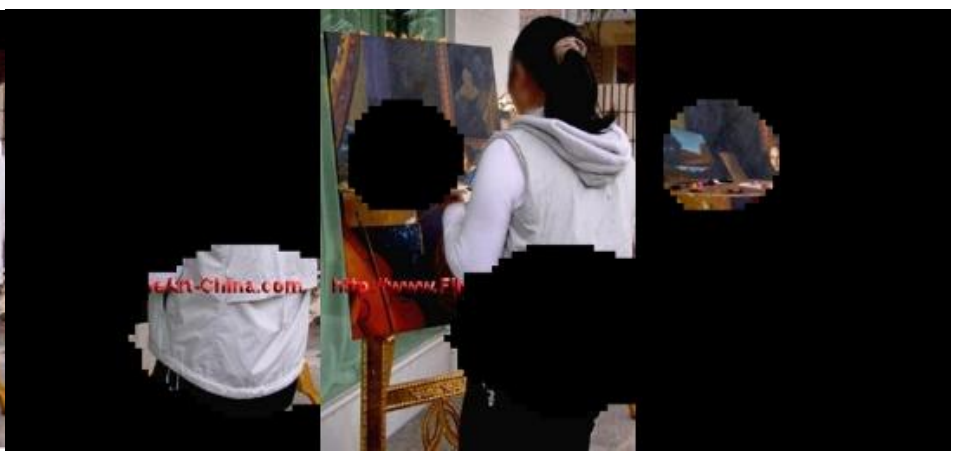

(b) GoogleNet

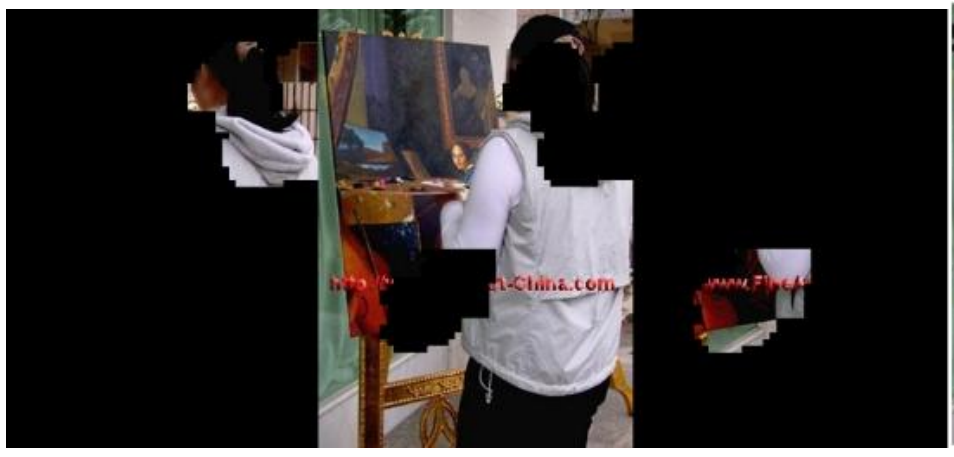

(c) ResNet-50

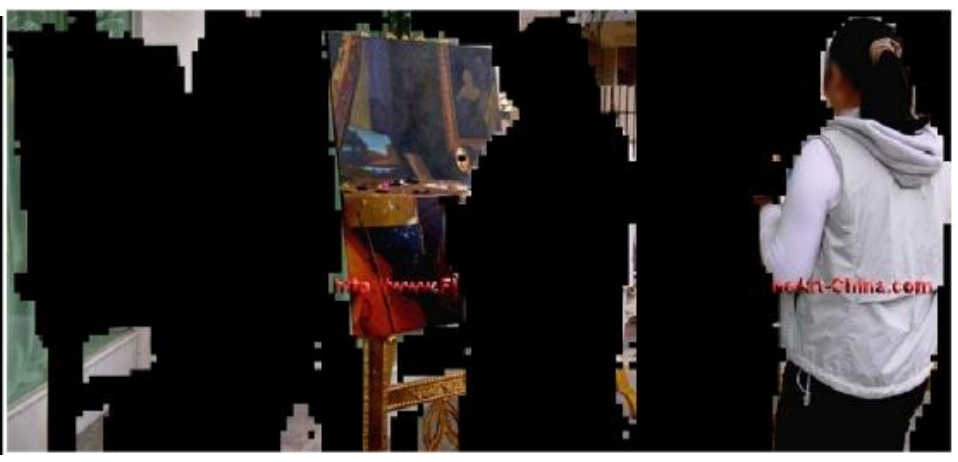

(d) SAM 2

Figure 7 Segmentation results

As shown in Figure 7, the segmentation results are consistent with the observations in Figure 6, suggesting that the PRM tensor derived from EfficientNet-B0 is better suited to discovering semantically meaningful regions, although SAM 2 produces smoother boundary for the person, the paint try is not segmented from the image.

### 3.3. Validation of the role of SCR profiles

Next, we examine whether SCR profiles can help identify promising PRM tensors across the evaluated pre-trained models and layer configurations. As an example, for the image shown in Figure 2, Figure 8 shows the SNR profiles of its PRM tensor derived from EfficientNet-B0, GoogleNet, ResNet-50, and SAM 2, respectively.

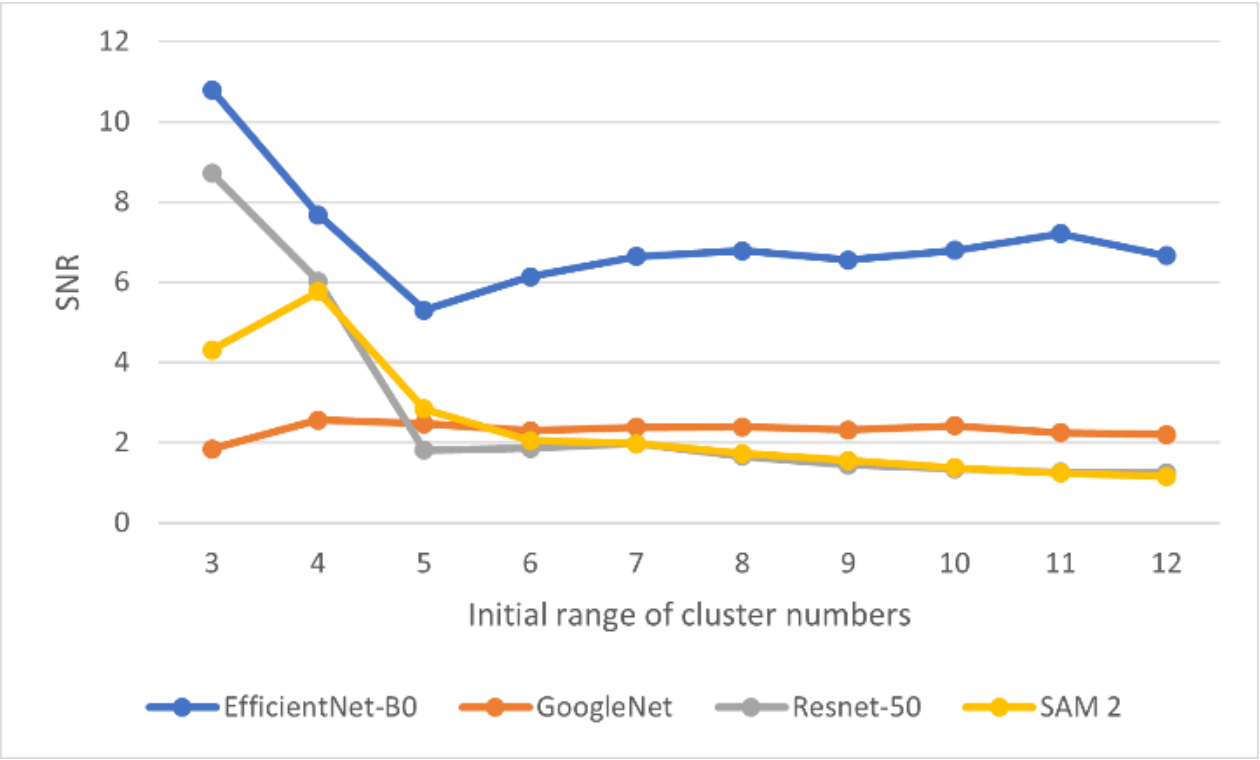

Figure 8 SNR profiles of PRM tensor from different configurations

As shown in Figure 8, the SNR profiles of the PRM tensors derived from EfficientNet-B0 consistently exhibit higher SNR values across the evaluated initial ranges of cluster numbers than those derived from the other pretrained models. In particular, when the threshold is set to 6, the number of SNR values exceeding this threshold is 8 for EfficientNet-B0, compared with only 1 for ResNet-50 and none for GoogLeNet and SAM 2. These results support the observations in Figures 6 and 7. Combined with the results presented in Figures 3, 4, and 5, these observations support the conjecture that PRM tensors exhibiting a higher proportion of relatively high SNR values over an initial range of cluster numbers tend to yield more semantically meaningful regions across multiple granularities.

Additional qualitative experiments on approximately 100 randomly selected images from general image collections and benchmark datasets showed similar observations. the EfficientNet-B0 PRM tensors tended to exhibit higher proportion of relatively high SNR values over the initial range of cluster numbers than the other models, and to yield more semantically meaningful regions. These findings suggest that SNR-profile analysis can help assess the suitability of PRM tensors for discovering semantically meaningful regions at different granularities through pixel-wise clustering without task-specific training.

### 3.4 Ablation study

To further investigate the effectiveness of PRM tensor in clustering semantically meaningful regions, we conduct an ablation study comparing PRM tensor with the original channel-wise feature maps without SVD. The channel-wise feature maps extracted from Layers A+B+C of EfficientNet-B0 are sorted in descending order and arranged into a tensor that preserves the structural format of the PRM tensor. As an example, using the image shown in Figure 2, Figure 9 presents the SNR profiles obtained under these two conditions. Figure 10 shows the corresponding segmentation results.

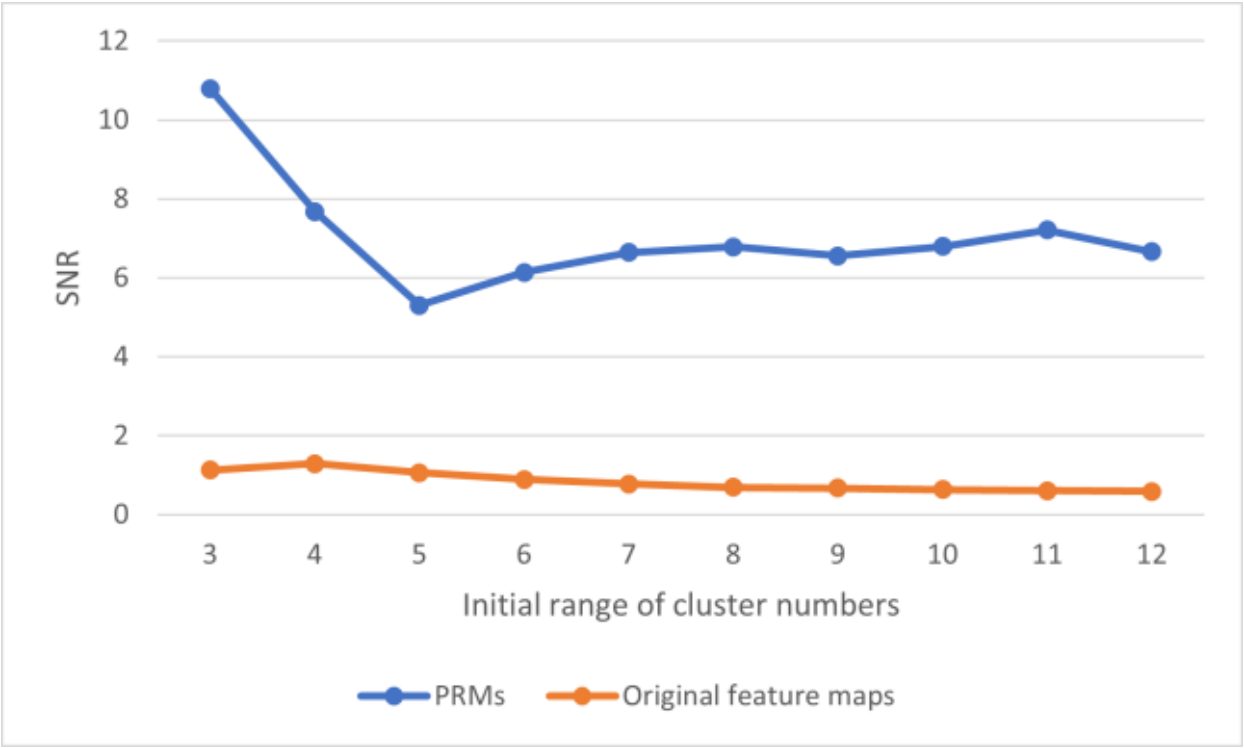

Figure 9 SNR profiles

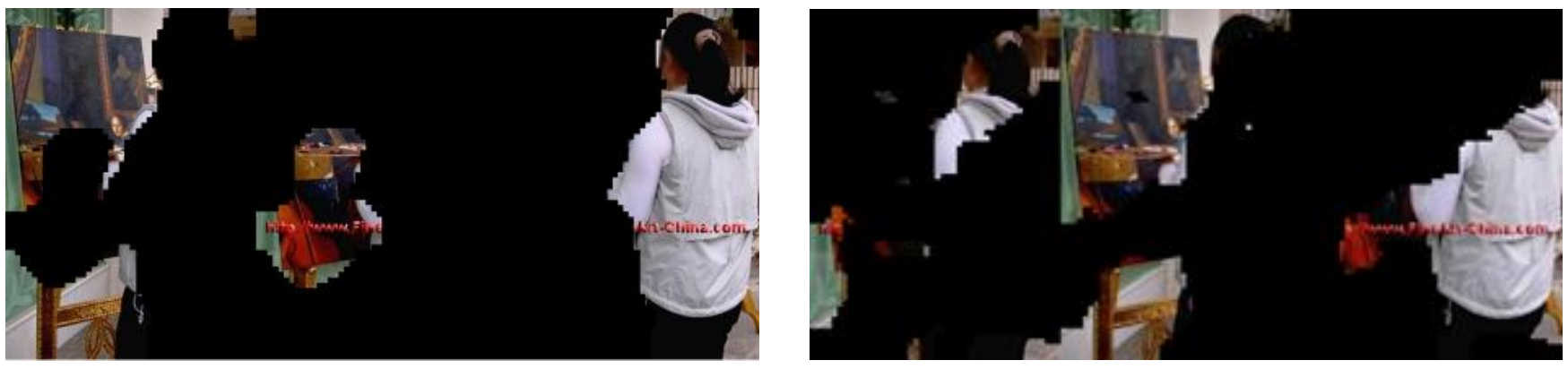

(a) By PRM tensor (b) By original feature maps

Figure 10 Segmentation results

As shown in Figure 9, the PRM tensor obtained through SVD consistently yield much higher SNR values than the original feature maps. This improvement is also reflected in Figure 10, where the PRM tensor produce higher-quality and more semantically meaningful regions than the original feature maps. These results show the contribution of SVD to enhancing the suitability of pretrained visual representations for discovering semantic granularity through clustering.

Next, to validate the effectiveness of setting the pixel-wise representation dimensionality of PRM tensors to the target number of clusters, we conduct an ablation study comparing the SNR profiles obtained under three different settings: $n_p = k$, $n_p = n_e$ equal to the slowdown point, and $n_p = 12$. Here, the slowdown point denotes the position of the first slowdown point in the singular value spectrum. For this example, $n_e = 3$. Using the image shown in Figure 2 as an example, Figure 11 presents the SNR profiles of the PRM tensor derived from EfficientNet-B0 under these three settings. Figure 12 shows the corresponding segmentation results for k=4.

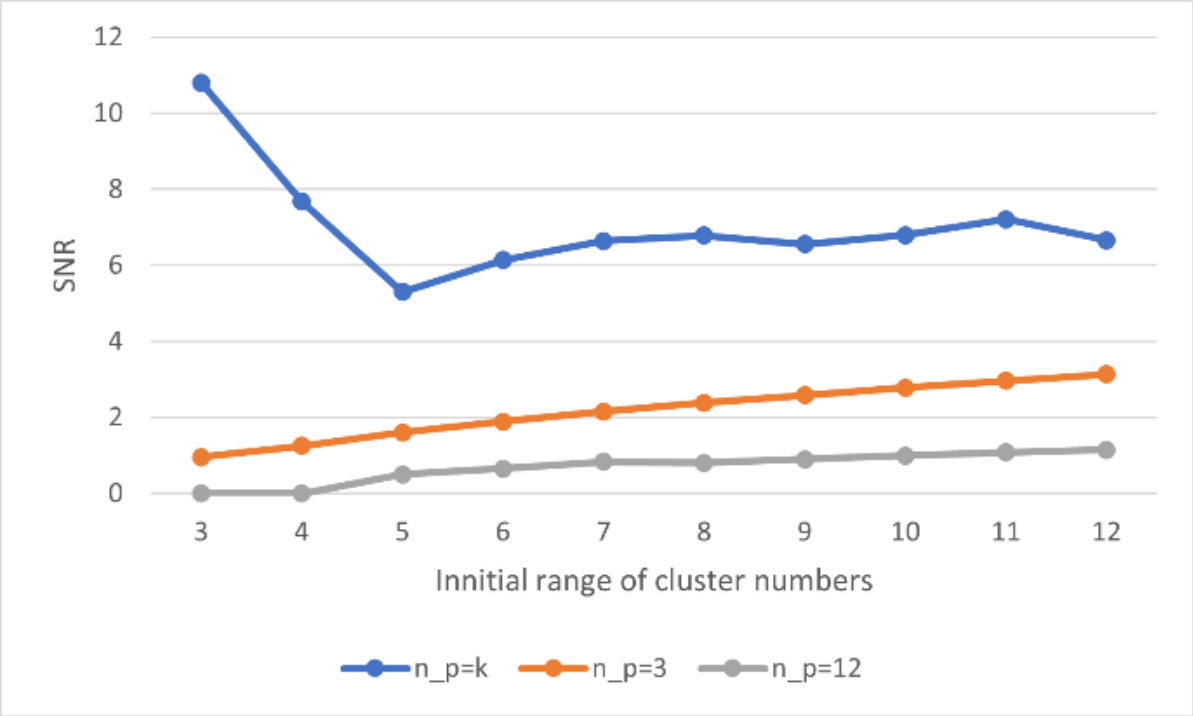

Figure 11 SNR profiles

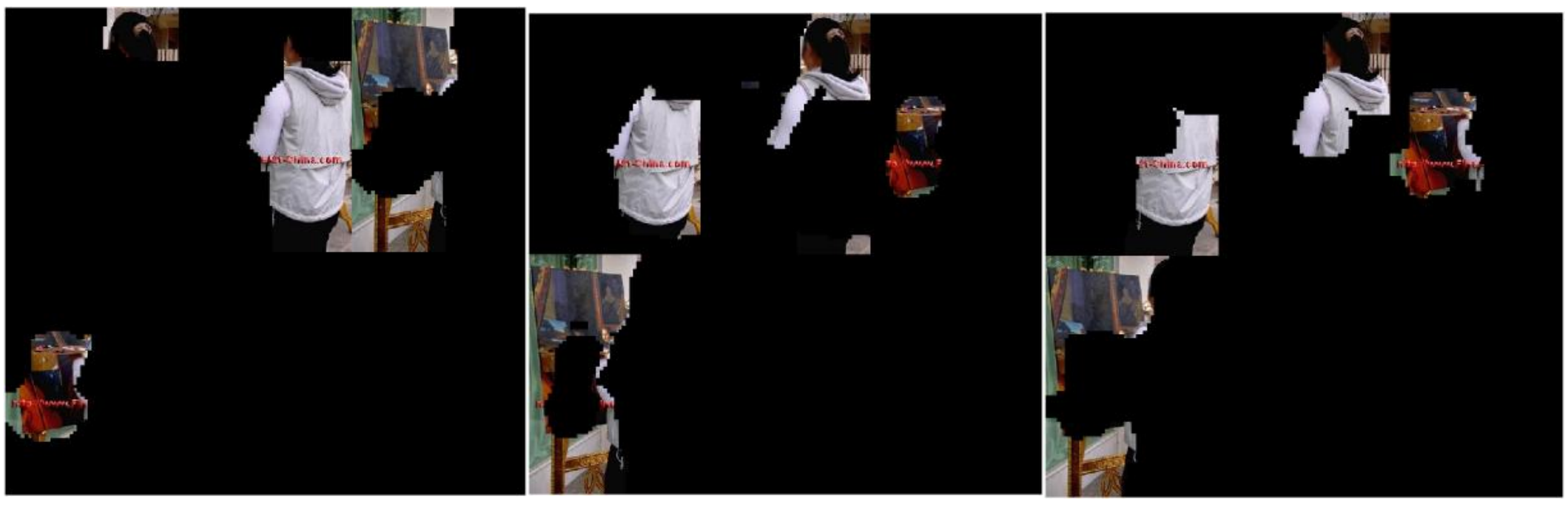

(aa) $n_p = k$ (b) $n_p = n_e$ $n_p = 12$

Figure 12 Segmentation results

As shown in Figure 11, the SNR values obtained with $n_p = k$ are much higher than those obtained with the other two settings. This improvement is also reflected in Figure 12, where $n_p = k$ produces higher-quality and more semantically meaningful regions than the alternative settings. These results show the effectiveness of matching the pixel-wise representation dimensionality to the target number of clusters in improving clustering performance when using a suitable PRM tensor.

### 3.5 Performance evaluation on benchmark datasets

To evaluate the feasibility of the proposed method, we employ Hungarian-matched intersection over union (HIoU) to assess the segmentation performance of clustered regions across multiple semantic granularities. HIoU evaluates class-agnostic segmentation by optimally matching predicted and ground-truth masks using the Hungarian algorithm. It builds an IoU matrix, finds one-to-one assignments maximizing total overlap, and computes mean IoU over matched pairs, ensuring fair, label-free comparison. The HIoU expression is usually written as:

$$HIoU = \frac{1}{N}\sum_{i=1}^{N}\frac{|P_i \cap G_{\pi(i)}|}{|P_i \cup G_{\pi(i)}|} \qquad (7)$$

Where, $P_i$ denotes predicted mask $i$, $G_{\pi(i)}$ denotes ground-truth mask matched to $P_i$, and $\pi$ is the optimal one-to-one assignment determined by the Hungarian algorithm.

Figure 13 presents the mean HIoU (mHIoU) results on the VOC-2012 validation dataset using different EfficientNet-B0 layer configurations under adaptive and GT-based cluster number settings. The results show that integrating multiple feature layers consistently improves segmentation performance under both settings, with Layers A+B+C achieving the highest mHIoU. This observation is consistent with the findings presented in Figure 3, further supporting the usefulness of SCR-profile analysis as an indicator for selecting suitable feature combinations.

Notably, the adaptive setting yields mHIoU values only about 2% lower than those of the GT-based setting across all configurations. This relatively small performance gap suggests that the proposed adaptive strategy can identify suitable cluster numbers without prior knowledge of the number of GT masks while maintaining competitive segmentation quality. These observations provide further evidence of the feasibility of the proposed method for discovering semantically meaningful regions at multiple granularities.

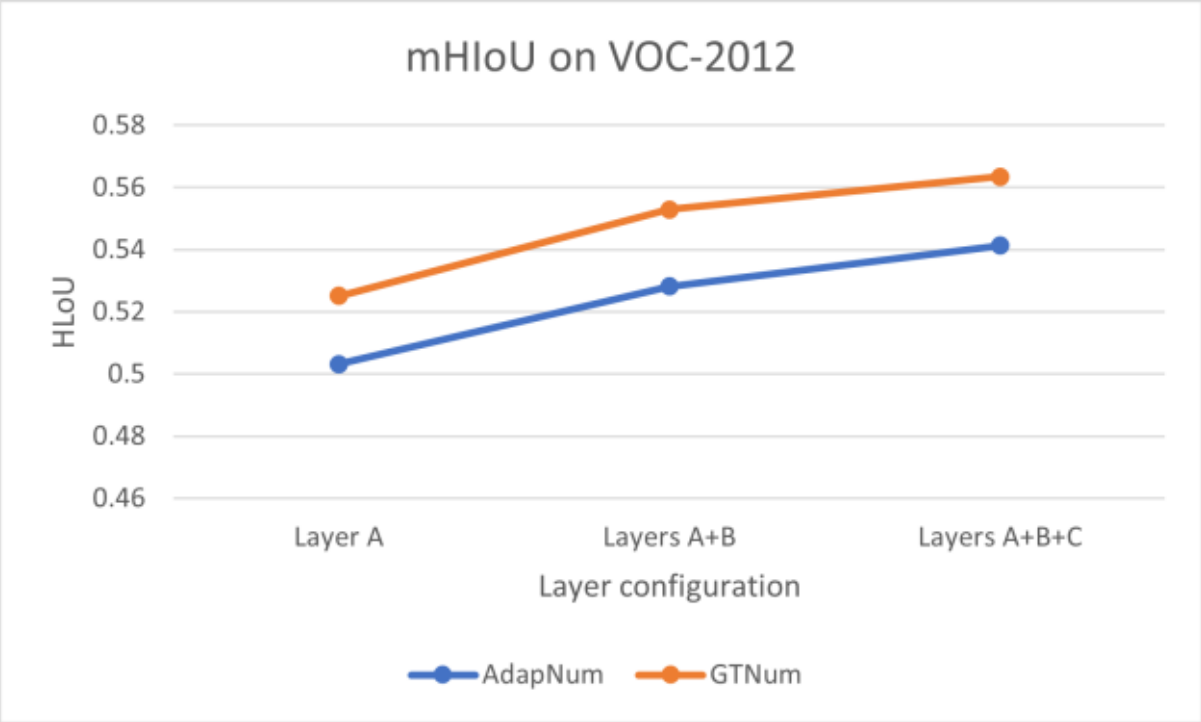

Figure 13 HIoU on VOC-2012

Furthermore, we compare the segmentation performance of the proposed approach with that of state-of-the-art (SOTA) methods on the VOC-2012 and COCO-2017 validation datasets. Table 1 summarizes the mHIoU results reported in previous studies and obtained using our approach. For COCO, N is defined as the number of predicted segments minus one, because stuff regions are not included in the GT annotations. “ANum” denotes the proposed adaptive setting, whereas “GTNum” denotes a reference setting in which the number of clusters is set to the number of GT masks.

Table 1 Comparison to SOTA results of mHIoU (%)

| Method | VOC-2012 | COCO-2017 |
|---|---|---|
| Jiao [1] | 59.8 | |
| DiffCut [3] | **65.2** | 34.1 |
| DiffSeg [5] | 48.2 | 31.7 |
| MaskCut [6] | 53.8 | 30.1 |
| Kim [13] | 53.3 | 21.2 |
| Simeoni [20] | 42.9 | |
| Dai (ours, ANum) | 54.1 | 32.5 |
| Dai (ours, GTNum) | 56.3 | **34.9** |

The results show that the proposed approach achieves competitive performance on both VOC-2012 and COCO-2017. On VOC-2012, the GTnum configuration obtains 56.3% mHIoU, outperforming several existing methods, including Kim, DiffSeg, and MaskCut, although it does not surpass Jiao or DiffCut. On COCO-2017, GTnum achieves 34.9% mHIoU, the competitive score among the compared methods, exceeding DiffCut (34.1%). The ANum configuration also achieves competitive results, outperforming DiffSeg and MaskCut on both datasets. These results support the feasibility of discovering semantic granularities through clustering of pretrained visual representations without task-specific training.

### 3.6 Example discussion

Figure 14 presents the segmentation results, PRMs, and SNR profile for an image from the COCO-2017 validation set at different cluster numbers, using PRMs derived from Layers A+B+C of EfficientNet-B0 and embeddings extracted from SAM 2, respectively.

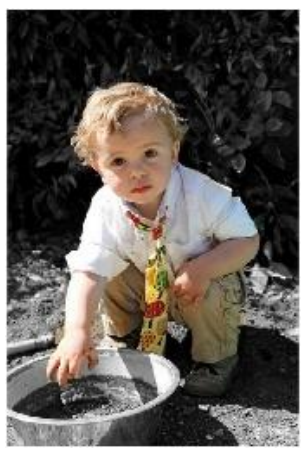

(a) Image A

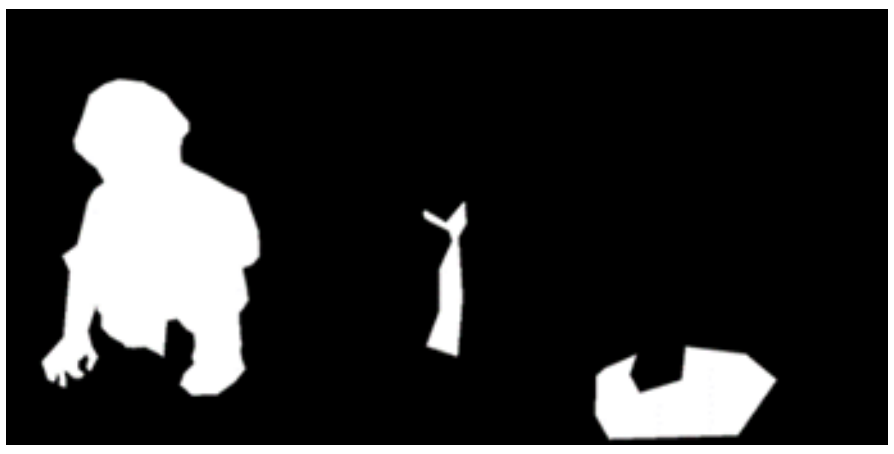

(b) GT masks

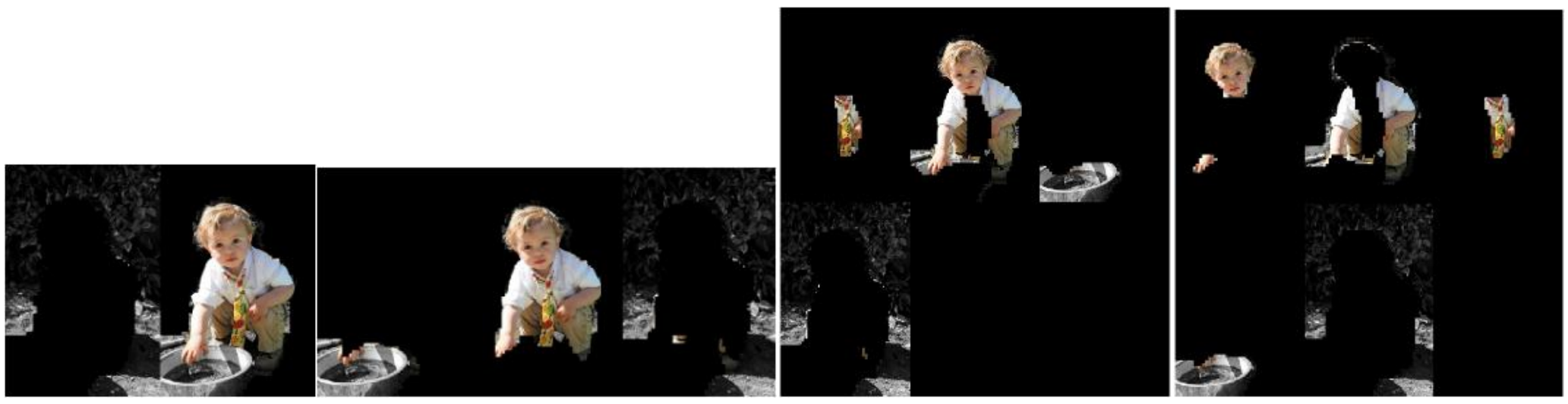

(c) Segments using EfficientNet-B0

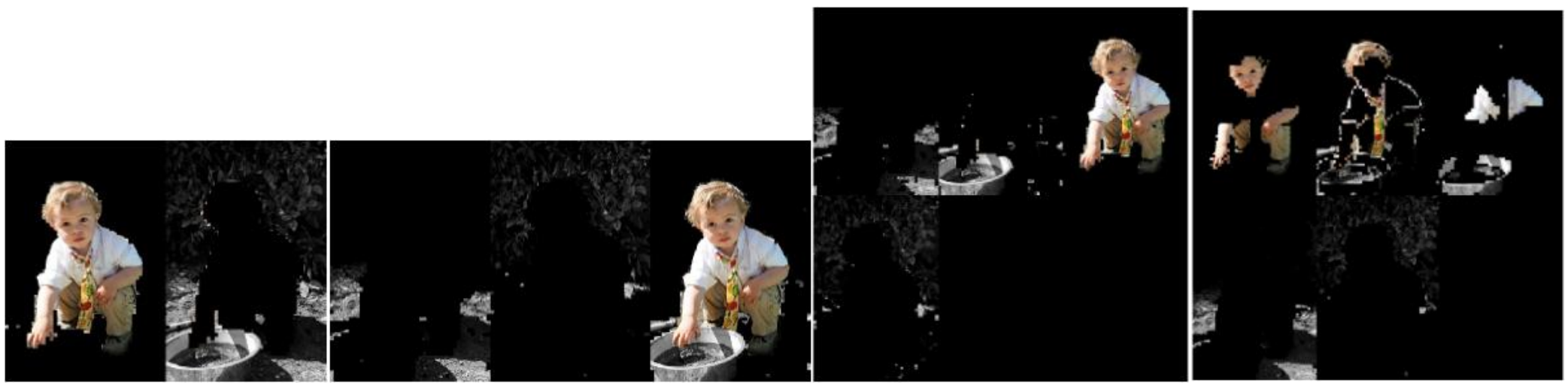

(d) Segments using SAM 2

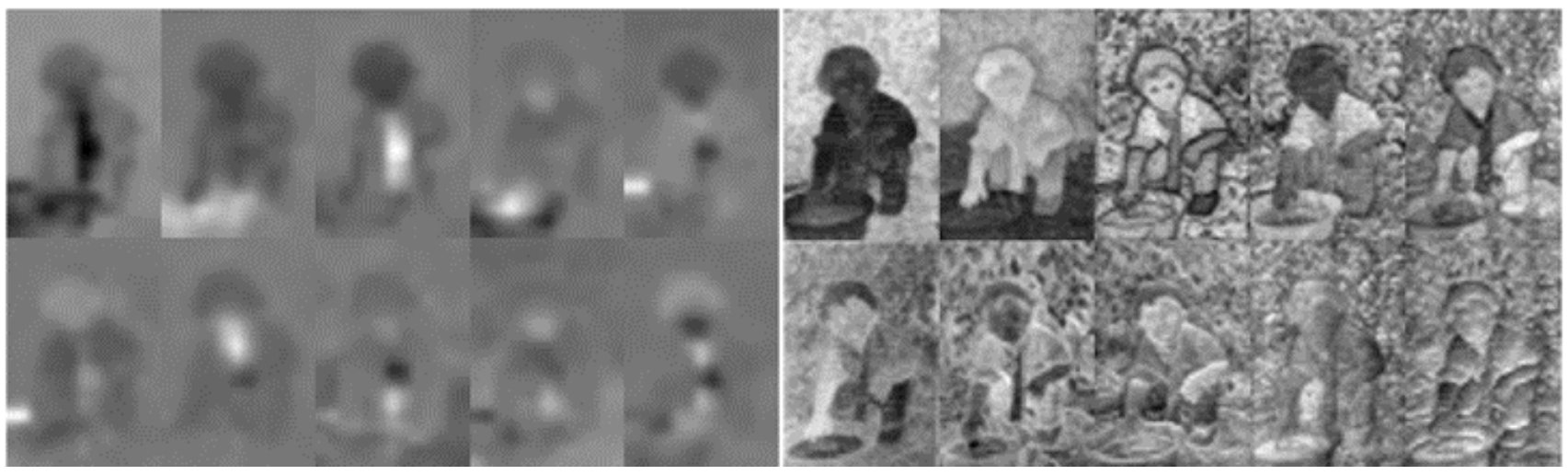

(e) PRMs of EfficientNetB0 (f) PRMs of SAM 2

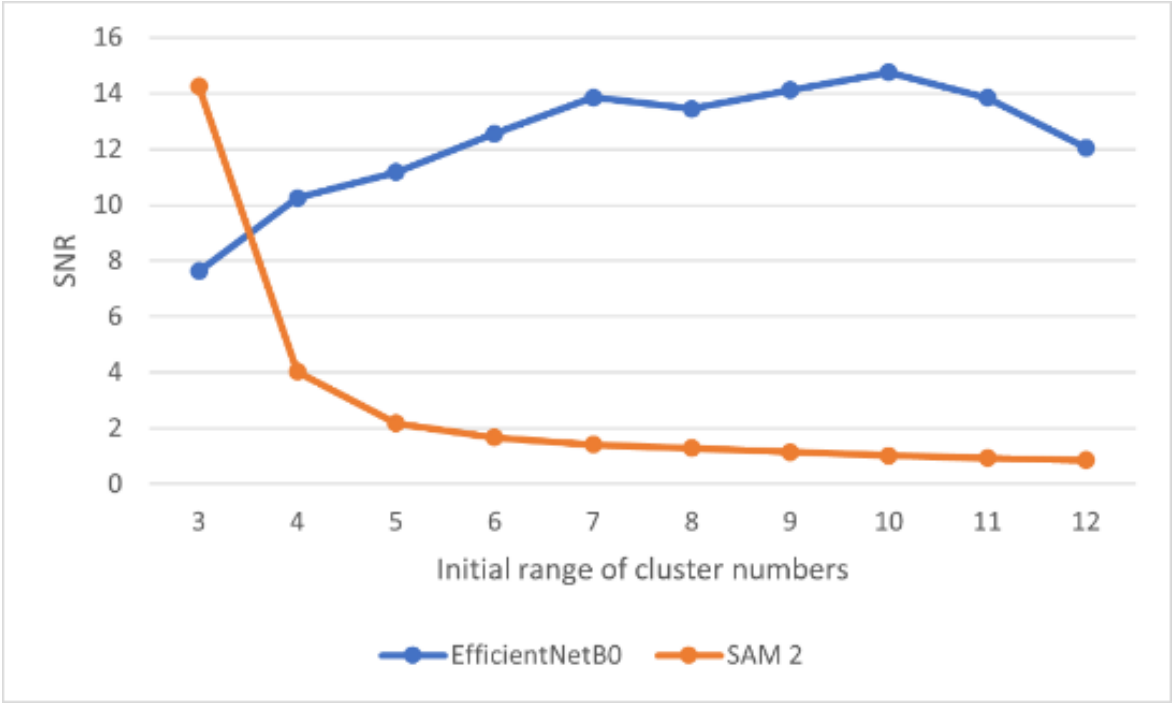

(g) SNR profiles

Figure 14 Results of Segmentation, PRMs, and SNR from image A

As shown in Figure 14(c), EfficientNet-B0 produces semantically meaningful segments across multiple granularities. When k=2, the child and the basin are separated from the background; when k=3, the child and the basin are segmented into distinct regions; when k=4, the necktie is further separated from the child; and when k=5, the head is further separated from the remaining body region of the child. According to the GT masks shown in Figure 14(b), the mHIoU is 0.7224 when k=3, which is adaptively determined, and 0.6052 when k=4, corresponding to the number of GT masks plus one.

In contrast, as shown in Figure 14(d), although SAM 2 successfully segments the child from the background when k=2, the child and the basin are assigned to the same segment when k=3. When k>4, the quality of the resulting object-level partitions deteriorates. These observations indicate that SAM 2 cannot consistently produce high quality partitions at granularity levels beyond k=3.

The PRMs shown in Figures 14(e) and 14(f), together with their corresponding SNR profiles in Figure 14(g), provide further insight into these differences. In particular, when the threshold is set to 6, the number of SNR values exceeding this threshold is 10 for EfficientNet-B0, compared with only 1 for SAM 2. For SAM 2, the higher SNR values in the initial range up to K=3 can be attributed to the clear separation of the child from other entities when the number of clusters is set to k=2.

Figure 15 presents the results for an image from the VOC-2012 validation dataset with the same experimental configuration as Figure 14.

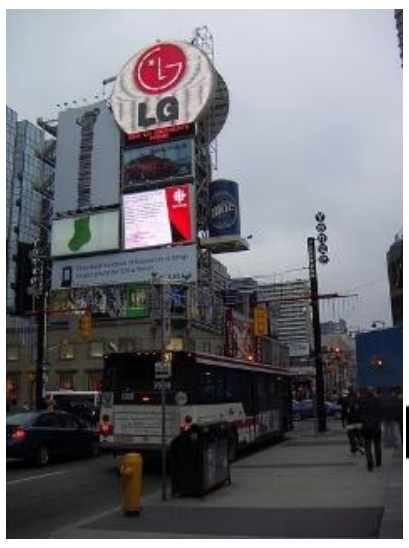

(a) Image B

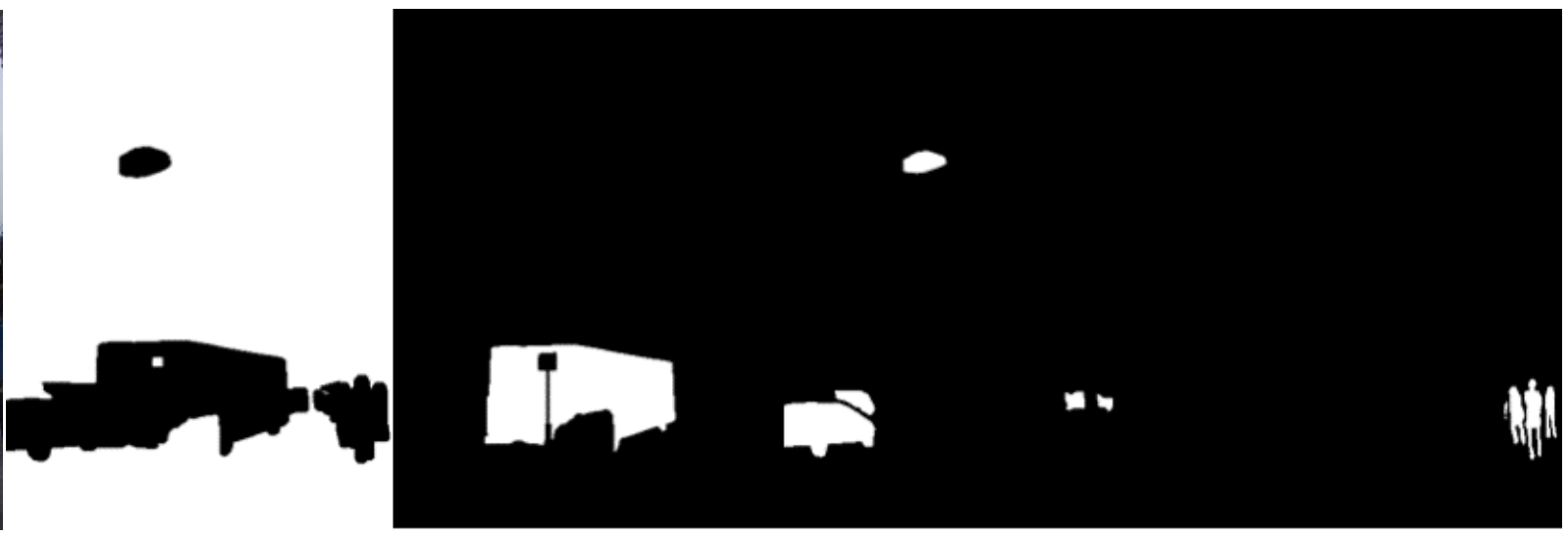

(b) GT masks

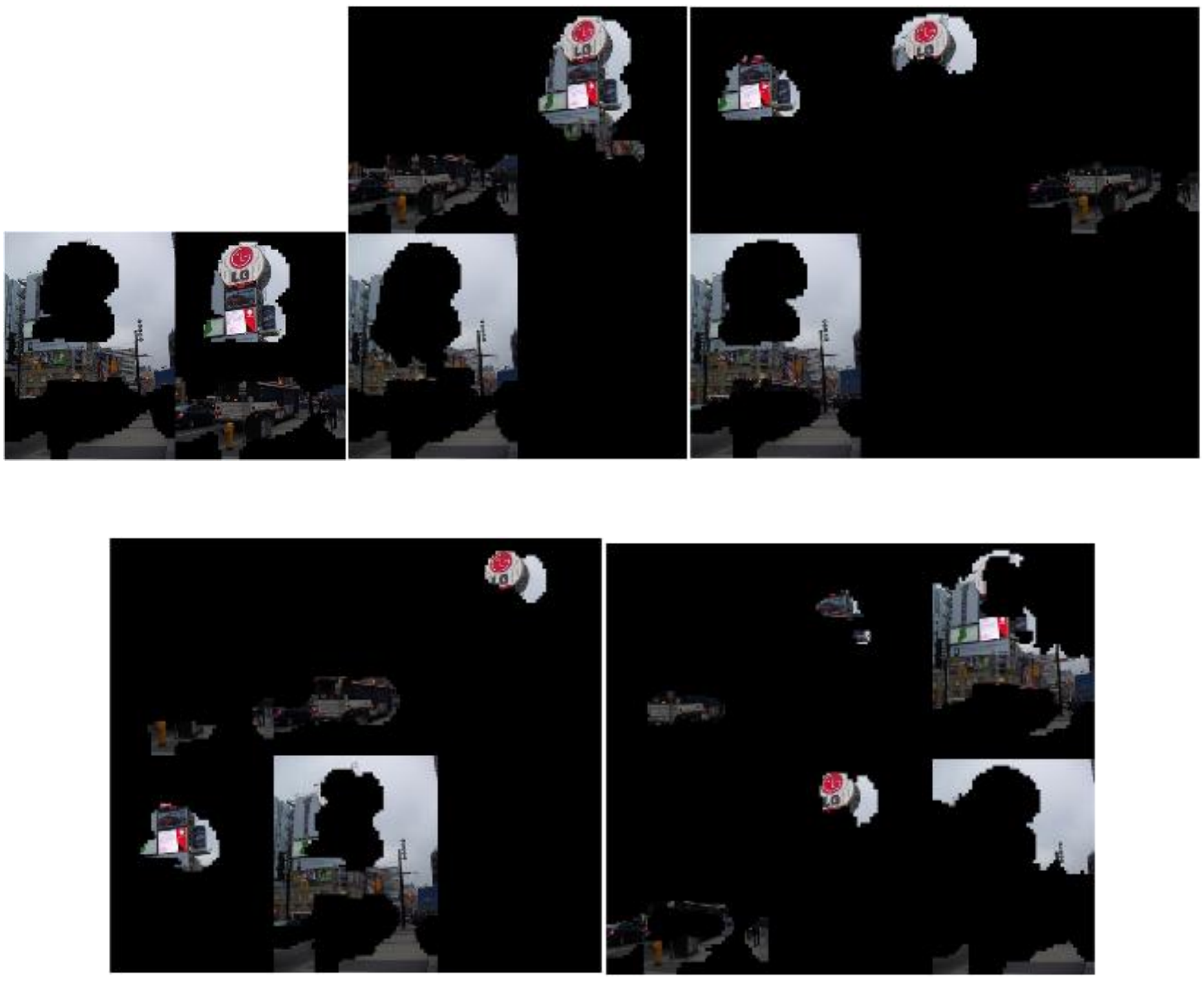

(c) Segments using EfficientNet-B0

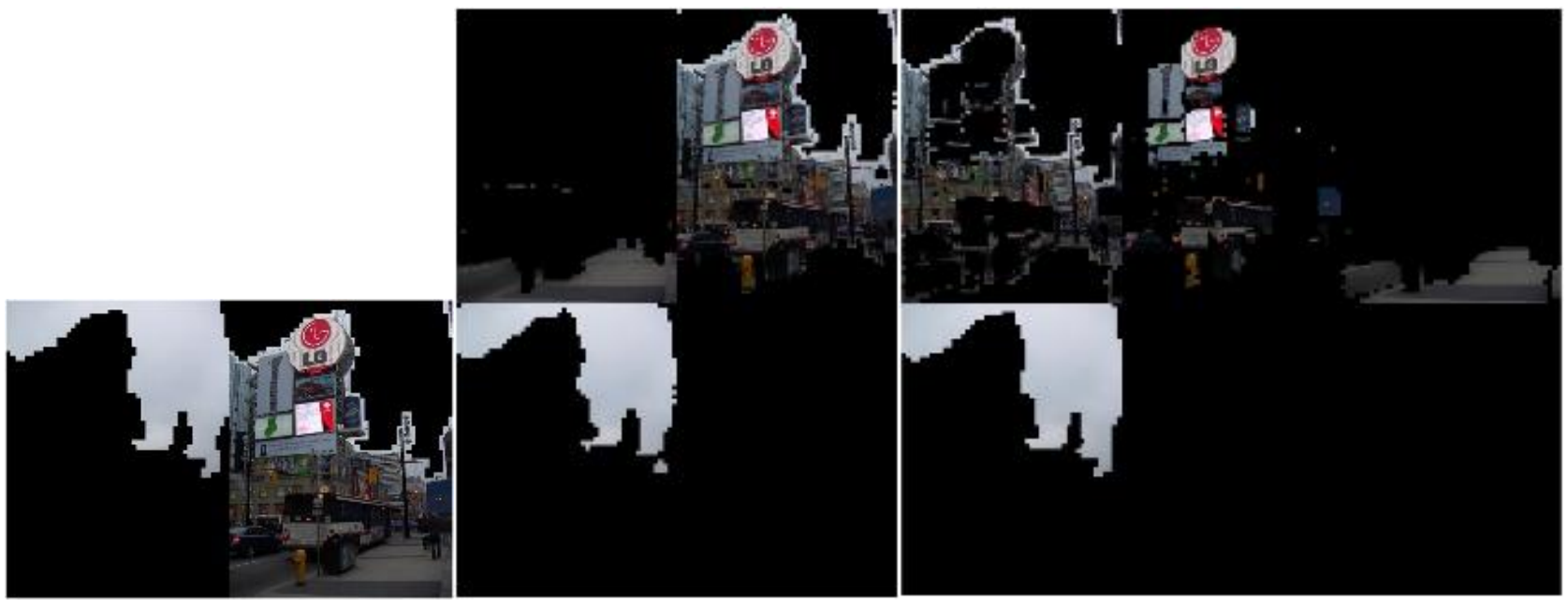

(d) Segments using SAM 2

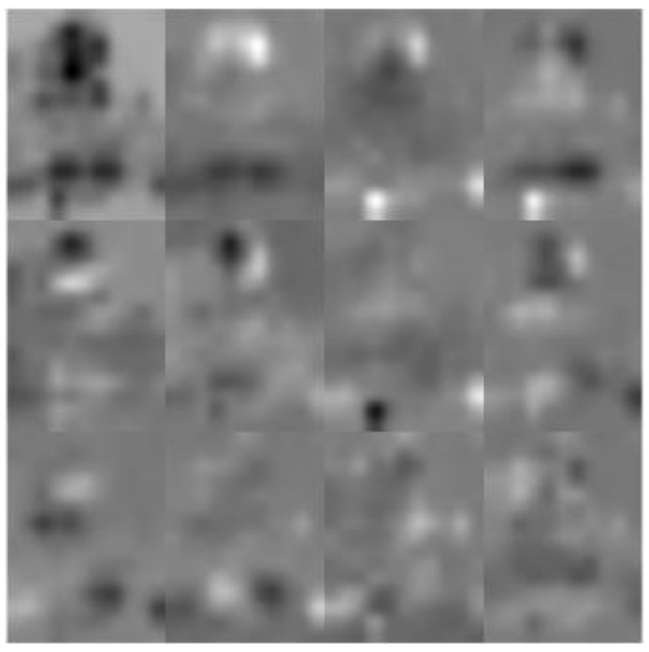

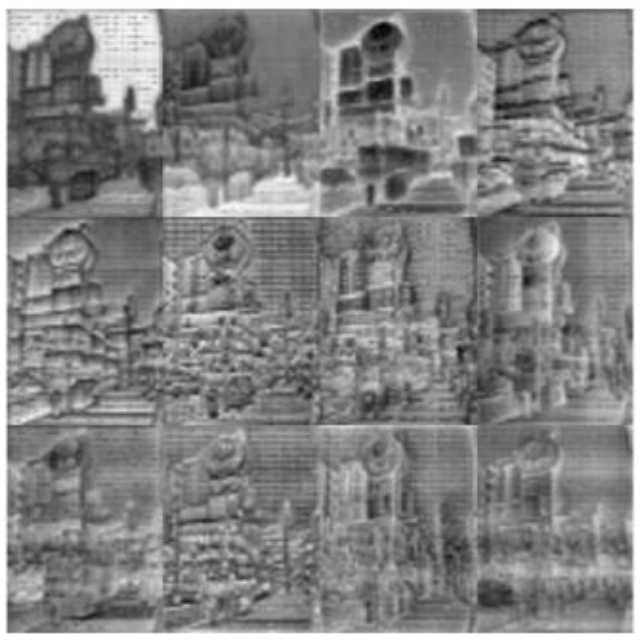

(e) PRMs of EfficientNetB0 (f) PRMs of SAM 2

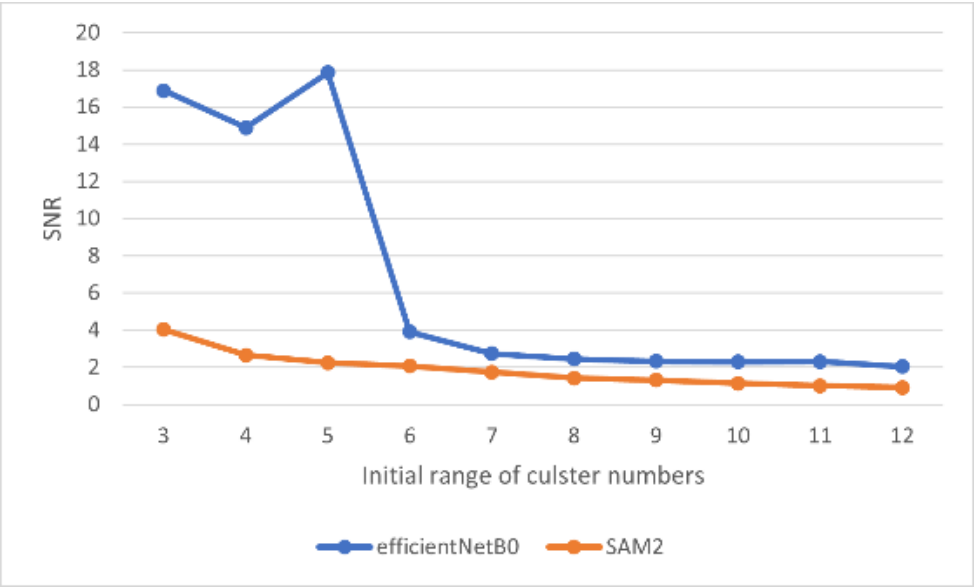

(g) SNR profiles

Figure 15 Results of Segmentation, PRMs, and SNR from image B

As shown in Figure 15(c), EfficientNet-B0 produces semantically meaningful segments across multiple granularities. When k=2, vehicles, billboards, and people are separated from the background. When k=3, the billboards are separated from the other objects. When k=4, the billboards are further subdivided into distinct regions. When k=5, the people are further separated, whereas when k=6, the quality of the resulting object-level partitions deteriorates. According to the GT masks shown in Figure 15(b), the mHIoU is 0.469 when k=2, which is adaptively determined, compared with 0.294 when k=4, corresponding to the number of GT masks. The GT masks include only vehicles and people, whereas the proposed method also segments billboards. Consequently, the mHIoU is relatively low, despite the clear segmentation of the billboards.

In contrast, as shown in Figure 15(d), SAM 2 fails to produce consistently high-quality segmentation results across the different cluster numbers.

The PRMs shown in Figures 15(e) and 15(f), together with their corresponding SNR profiles in Figure 15(g), provide further insight into these differences. In particular, when the SNR threshold is set to 6, three SNR values exceed this threshold for EfficientNet-B0, whereas none do for SAM 2.

Figure 16 shows the results for an additional real-world image outside the benchmark datasets, using the same experimental configuration as in Figure 14.

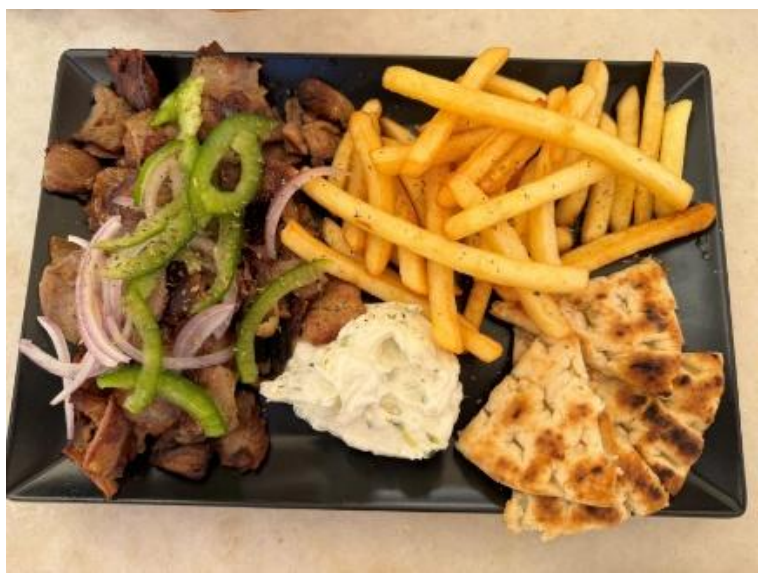

(a) Image C

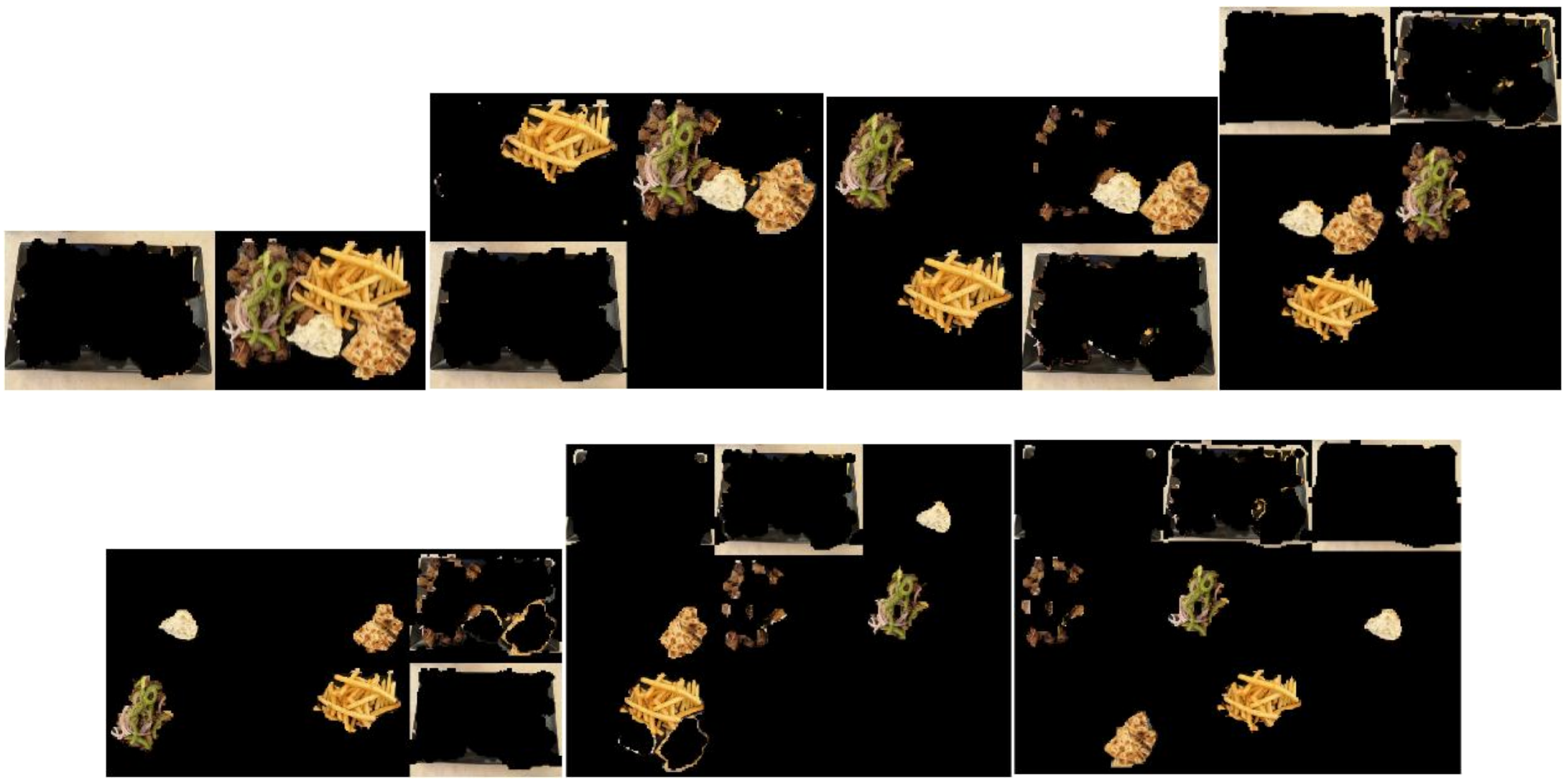

(b) Segments using EfficientNet-B0

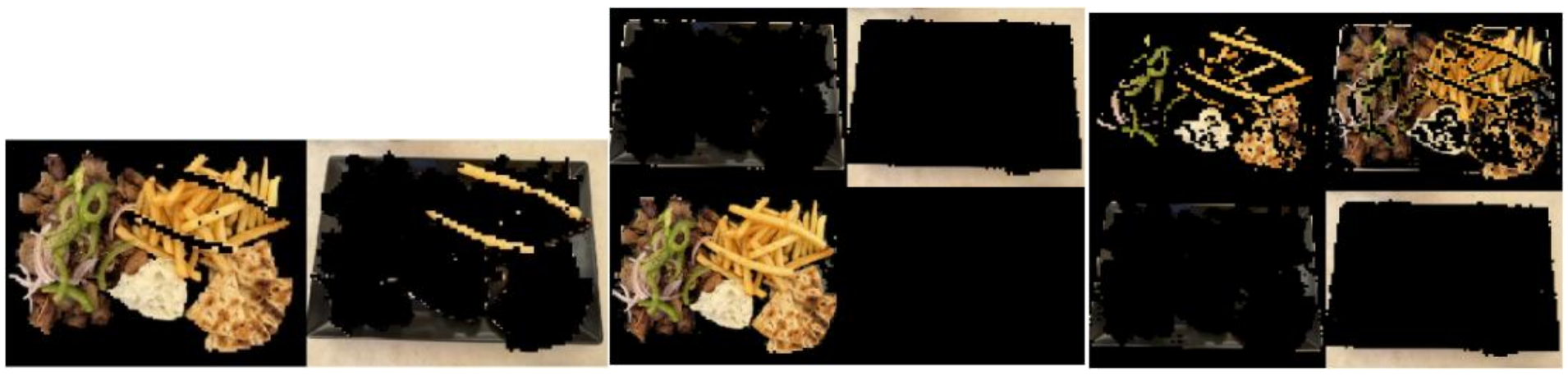

(c) Segments using SAM 2

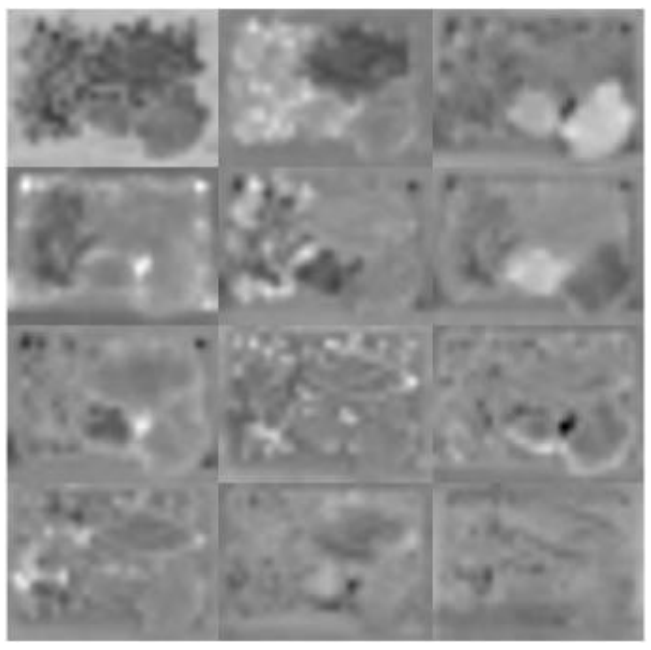

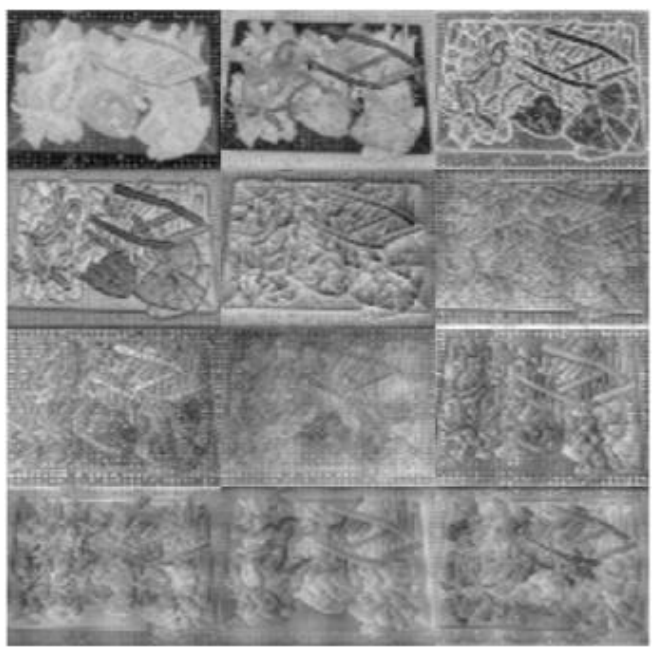

(d) PRMs of EfficientNetB0 (e) PRMs of SAM 2

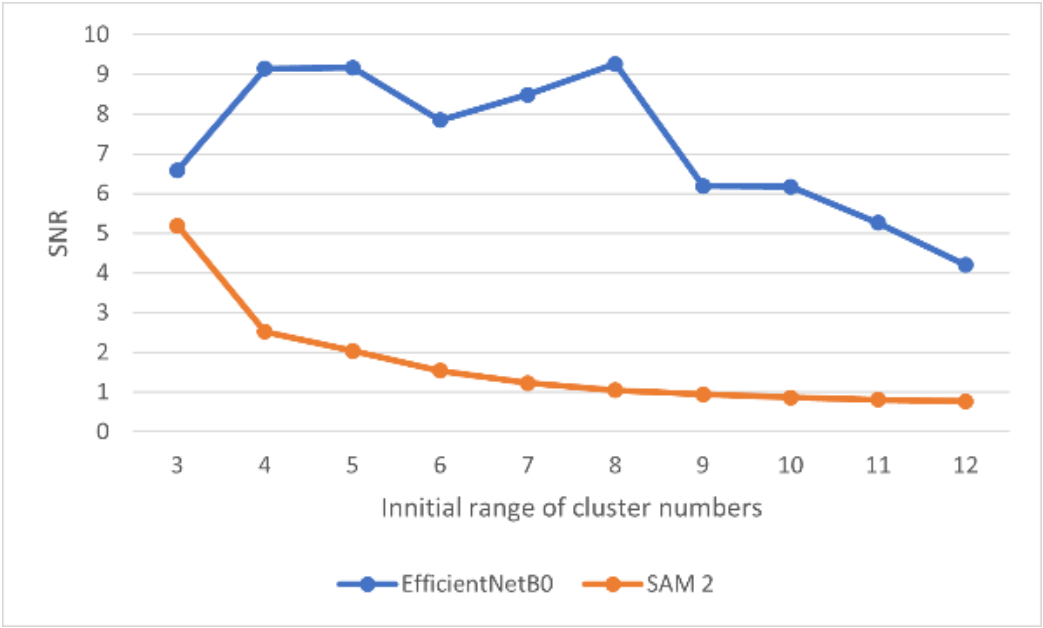

(f) SNR profiles

Figure 16 Results of Segmentation, PRMs, and SNR from image C

As shown in Figure 16(b), the ingredients in the image are progressively separated into distinct segments as the number of clusters increases. In contrast, as shown in Figure 16(c), SAM 2 fails to produce consistently high-quality segmentation results across different cluster numbers.

The PRMs shown in Figures 16(d) and 16(e), together with their corresponding SNR profiles in Figure 16(f), provide further insight into these differences. In particular, when the SNR threshold is set to 6, eight SNR values exceed this threshold for EfficientNet-B0, whereas none do for SAM 2.

Overall, these observations support the conjecture that PRM tensors exhibiting a higher proportion of relatively high SNR values over an initial range of cluster numbers tend to yield more semantically meaningful regions across multiple granularities. Furthermore, once a suitable PRM tensor has been identified through SNR-profile analysis, pixel-wise hierarchical clustering enables the level of semantic granularity to be controlled by varying the number of clusters, with the pixel-wise representation dimensionality of the PRM tensor set equal to the number of clusters.

## 4. Conclusion

This paper presented a representation-driven approach for discovering semantic granularities in images through clustering of pretrained visual representations without task-specific training. We proposed PRM tensor, constructed by concatenating feature maps from appropriately deep layers of the pretrained visual model and apply the singular value decomposition (SVD)-based projection. The experimental results showed that this projection captures the latent semantic-like spatial structures embedded in pretrained representations, making them accessible to subsequent semantic clustering.

We also introduced the SCR and an SCR-profile-based SNR as a proposed indicator for assessing the suitability of PRM tensors for semantic granularity discovery. Qualitative results indicated that PRM tensors exhibiting a higher proportion of relatively high SNR values over an initial range of cluster numbers tended to yield more semantically meaningful regions across multiple levels of granularity. Among the four evaluated pretrained models, EfficientNet-B0 provided the most suitable PRM tensors for this purpose. Furthermore, pixel-wise hierarchical clustering allowed the semantic granularity to be controlled by varying the number of clusters, with the pixel-wise representation dimensionality of the PRM tensor set equal to the number of clusters.

Furthermore, evaluations on the VOC-2012 and COCO-2017 validation datasets demonstrated competitive segmentation performance. The results obtained using the adaptive cluster-number configuration further suggested the feasibility of discovering meaningful image regions without task-specific training.

Future work will investigate methods for recognizing the resulting regions at multiple levels of granularity to facilitate image scene understanding.